\documentclass[sigconf]{acmart}

\definecolor{lightblue}{RGB}{230,255,255}
\usepackage{threeparttable}
\usepackage{multirow}
\usepackage{xcolor}
\usepackage{colortbl}
\usepackage{amsmath}
\usepackage{graphicx,subfig,caption}

\usepackage{amssymb}
\usepackage{etoolbox}
\AtBeginDocument{%
  }

\setcopyright{acmlicensed}
\copyrightyear{2025}
\acmYear{2025}
\acmDOI{XXXXXXX.XXXXXXX}

\begin{document}

\title{HeLo$:$ Heterogeneous Multi-Modal Fusion with Label Correlation for Emotion Distribution Learning}

\author{Chuhang Zheng}
\orcid{0009-0008-0451-5091}
\affiliation{%
  \department{College of Artificial Intelligence}
  \institution{Nanjing University of Aeronautics and Astronautics}
  \city{Nanjing}
  \country{China}
}
\email{h.zheng@nuaa.edu.cn}

\author{Chunwei Tian}
\orcid{0000-0002-6058-5077}
\affiliation{%
  \department{School of Computer Science and Technology}
  \institution{Harbin Institute of Technology}
  \city{Harbin}
  \country{China}
}
\email{chunweitian@163.com}

\author{Jie Wen}
\orcid{0000-0001-9554-2379}
\affiliation{%
  \department{School of Computer Science and Technology}
  \institution{Harbin Institute of Technology, Shenzhen}
  \city{Shenzhen}
  \country{China}
}
\email{jiewen_pr@126.com}

\author{Daoqiang Zhang}
\orcid{0000-0002-5658-7643}
\affiliation{%
  \department{College of Artificial Intelligence}
  \institution{Nanjing University of Aeronautics and Astronautics}
  \city{Nanjing}
  \country{China}
}
\email{dqzhang@nuaa.edu.cn}

\author{Qi Zhu}
\orcid{0000-0001-7740-292X}
\authornote{Corresponding author.}
\affiliation{%
  \department{College of Artificial Intelligence}
  \institution{Nanjing University of Aeronautics and Astronautics}
  \city{Nanjing}
  \country{China}
}
\email{zhuqi@nuaa.edu.cn}

\renewcommand{\shortauthors}{Chuhang Zheng, Chunwei Tian, Jie Wen, Daoqiang Zhang, $\&$ Qi Zhu}

\begin{abstract}
  Multi-modal emotion recognition has garnered increasing attention as it plays a significant role in human-computer interaction (HCI) in recent years. Since different discrete emotions may exist at the same time, compared with single-class emotion recognition, emotion distribution learning (EDL) that identifies a mixture of basic emotions has gradually emerged as a trend. However, existing EDL methods face challenges in mining the heterogeneity among multiple modalities. Besides, rich semantic correlations across arbitrary basic emotions are not fully exploited. In this paper, we propose a multi-modal emotion distribution learning framework, named \textbf{HeLo}, aimed at fully exploring the heterogeneity and complementary information in multi-modal emotional data and label correlation within mixed basic emotions. Specifically, we first adopt cross-attention to effectively fuse the physiological data. Then, an optimal transport (OT)-based heterogeneity mining module is devised to mine the interaction and heterogeneity between the physiological and behavioral representations. To facilitate label correlation learning, we introduce a learnable label embedding optimized by correlation matrix alignment. Finally, the learnable label embeddings and label correlation matrices are integrated with the multi-modal representations through a novel label correlation-driven cross-attention mechanism for accurate emotion distribution learning. Experimental results on two publicly available datasets demonstrate the superiority of our proposed method in emotion distribution learning.
\end{abstract}




\maketitle

\section{Introduction}
Emotion recognition plays a pivotal role in enhancing human-computer interaction (HCI) \cite{shneiderman2010designing,li2023effective,can2023approaches}, offering substantial benefits across various fields such as healthcare, education, and customer service~\cite{alarcao2017emotions}. Researchers have investigated various behavioral signals, including facial expressions~\cite{verma2021automer} and voice signals~\cite{yi2020improving}, as well as physiological signals like electroencephalogram (EEG)~\cite{ye2022hierarchical}, electrocardiogram (ECG)~\cite{hsu2017automatic}, and electromyogram (EMG)~\cite{cheng2008emotion}, to detect emotional states. While many studies have concentrated on identifying basic emotions using the valence-arousal (VA)~\cite{10349925} or valence-arousal-dominance (VAD)~\cite{lang1995emotion} model, there is an increasing awareness of the complexity and mixed nature of human emotions. Research indicates that individuals may simultaneously experience multiple emotions with varying intensities at the same time \cite{williams2002can,zhao2020multi}. However, most existing emotion recognition methods tend to focus on single-class models \cite{jia2019facial,zhang2018text,zheng2016personalizing}, which may not fully capture the varied, complex, and sometimes ambiguous nature of human emotions. Accurate identification of these complex emotional states is essential, as it reflects more realistic human experiences and interactions, underscoring the need for advanced emotion recognition systems that can interpret these nuances effectively.

\begin{figure}[t]
\centering
\includegraphics[width=3.3in]{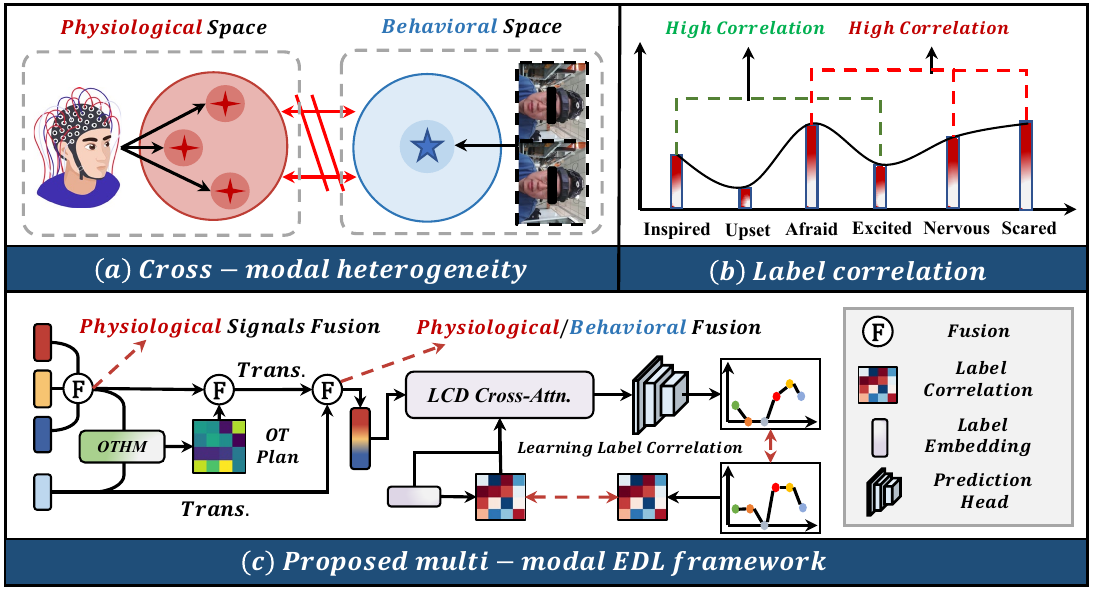}
\caption{\textbf{The challenges in emotion distribution learning and illustration of the proposed method.} (a) cross-modal heterogeneity, (b) label correlation, (c) the schematic diagram of the proposed framework.} \label{fig1}
\end{figure}

In contrast with multi-label emotion recognition \cite{li2019blended,zhang2020multi,zhang2021multi}, which typically classifies emotions into distinct and singular categories such as happiness, sadness, and anger, mixed emotion recognition aims to capture the complexity of coexisting emotions that do not easily fit into traditional labels. Furthermore, mixed emotion recognition evaluates the intensity of each emotion present, whereas multi-label approaches merely identify the presence or absence of each emotional state. Jia et al.~\cite{jia2019facial} adopted an emotion distribution learning method that exploits label correlations locally for facial expression recognition. Zhang et al.~\cite{zhang2018text} proposed a multi-task convolutional neural network for text emotion analysis. Existing single-modal methods struggle to fully utilize the complementary and correlative aspects between different modalities, overlooking the fact that emotions can be expressed through a combination of physiological responses and external manifestations. 

Researchers have explored the efficacy of integrating multi-modal data to enhance emotion recognition performance~\cite{10930808}. Peng et al.~\cite{peng2024carat} proposed CARAT to model fine-grained modality-to-label dependencies and exploit the modality complementarity by a shuffle-based aggregation strategy for multi-modal multi-label emotion recognition. Shu et al.~\cite{shu2022emotion} proposed an EDL convolutional neural network to predict the emotion distribution from the stacked physiological signals. Liu et al.~\cite{liu2023emotion} developed EmotionDict to learn a set of basic emotion elements in the latent feature space for emotion distribution learning. However, as the diagram shown in Figure \ref{fig1} (a), these methods ignore the intrinsic correlation and heterogeneity among different modalities, resulting in a sub-optimal recognition performance. What’s more, they cannot fully exploit the label correlation across different basic emotions. As can be seen from Figure \ref{fig1} (b), “afraid” and “scared” may be highly correlated. Leveraging the label correlations as a guidance can enhance the performance of emotion distribution learning.

To address the above limitations, in this paper, we propose a multi-modal emotion distribution learning framework, named HeLo, as depicted in Figure \ref{fig1} (c), which effectively mines the cross-modal heterogeneity within physiological and behavioral data, facilitating the learning of the label correlation to guide emotion distribution learning. In brief, the main contributions of our proposed model can be summarized as follows:
\begin{itemize}
    \item Due to differences in heterogeneity across modalities, we first adopt a cross-attention mechanism to fuse the physiological data. Then, an optimal transport (OT)-based heterogeneity mining module is devised to effectively fuse the physiological and behavioral representations.
    \item For the learning of label correlation, we introduce a learnable label embedding, which is constrained by its learnable label correlation and ground-truth label correlation. Furthermore, the learnable label embeddings and label correlation are integrated through a novel label correlation-driven cross-attention mechanism.
    \item Extensive experiments on two multi-modal emotion datasets demonstrate that our proposed model outperforms state-of-the-art models in both subject-dependent and subject-independent scenarios.
\end{itemize}

\section{Related Work}

\subsection{Label Distribution Learning}

In order to address the problem that both the labels and their intensities are necessarily needed, Geng et al. systematically proposed label distribution learning~\cite{geng2016label}. According to the proposed LDL strategies, traditional LDL methods can be divided into three categories, i.e., Problem Transformation (PT) \cite{geng2013label,geng2014facial}, Algorithm Adaptation (AA) \cite{geng2013facial,geng2010facial}, and Specialized Algorithms (SA) \cite{geng2016label,geng2013facial}. Recently, Wen et al. proposed the learning objectives and evaluation metrics for label distribution learning, namely Cumulative Absolute Distance (CAD), Quadratic Form Distance (QFD), and Cumulative Jensen-Shannon divergence (CJS). LDL-LRR~\cite{jia2023label} adopted a novel ranking loss function for label distribution characteristics and combined it with KL-divergence as the loss term of the objective function. Gao et al.~\cite{gao2017deep} proposed deep label distribution learning (DLDL), which utilizes label ambiguity in both feature learning and classifier learning to establish an end-to-end LDL framework. 

To improve the performance of LDL, researchers have made attempts to exploit the label correlations. Jia et al.~\cite{jia2019facial} employed a local low-rank structure to capture the label correlation implicitly to enhance facial expression recognition. Kou et al.~\cite{ijcai2024p478} proposed TLRLDL, which leverages low-rank label correlations specifically within an auxiliary multi-label learning framework. Different from these methods, we develop an end-to-end EDL framework, which makes full use of the complementarity between physiology and behavior signals, and alleviates the negative impact of discrete emotional ambiguity on EDL through label correlation learning.

\subsection{Muli-modal Emotion Recognition}

As multi-modal data provide a more comprehensive understanding of human emotional states \cite{ngai2022emotion,zheng2023prior,10938180}, multi-modal emotion recognition has emerged as an important topic in the field of affective computing. The dominant multi-modal emotion recognition methods can be divided into two subfields: single-label emotion recognition (SLER) and multi-label emotion recognition (MLER). SLER recognized emotions using a single emotion label for each discrete time. MAET~\cite{jiang2023multimodal} processed multi-modal inputs with specialized attention modules and alleviates subject discrepancy via adversarial training. While the goal of MLER is to assign an arbitrary number of emotion category labels to an input sample. Zhang et al.~\cite{zhang2020multi} integrated transformer and multi-head modality attention to effectively fuse text, visual, and audio data and predict emotions. Compared with SLER and MLER, emotion distribution learning (EDL) has gained increasing attention, as it evaluates the intensity of each basic emotion, which is crucial for a comprehensive representation of complex emotional space. EmotionDict~\cite{liu2023emotion} utilized behavior and physiological responses to offer complementary information for emotion recognition and disentangled emotion features of an emotional state into a weighted combination of a set of basic emotion elements. Shu et al.\cite{shu2022emotion} proposed EDLConv the combine multi-modal peripheral physiological signals for emotion distribution learning. In this paper, we propose a novel EDL framework for integrating multi-modal emotional data, and the model’s performance is improved with the guidance of learned label correlation.

\begin{figure*}[ht]
\centering
\includegraphics[width=\textwidth]{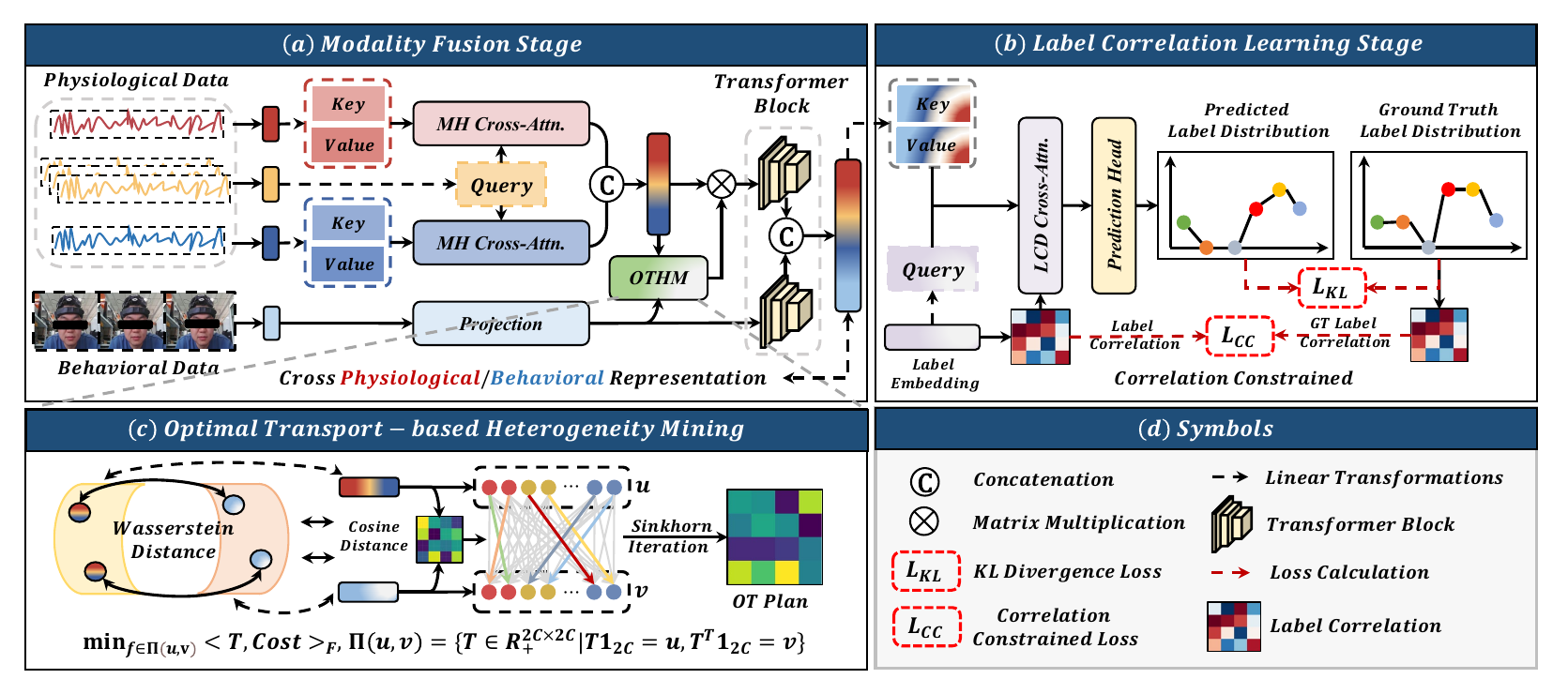}
\caption{\textbf{The pipeline of the proposed HeLo}. \textbf{(a) Modality fusion stage:} We first adopt a cross-attention mechanism to fuse the physiological data. Then the OT-based heterogeneity mining module is introduced to integrate the physiological and behavioral representations. \textbf{(b) Label correlation learning stage:} A learnable label embedding is introduced to explore the label correlation that can be combined with the multi-modal features through a novel label correlation-driven cross-attention mechanism.} \label{fig2}
\end{figure*}

\section{Methodology}

The overall framework of the proposed method is illustrated in Figure \ref{fig2}, which consists of physiological signals fusion, OT-based heterogeneity mining, and label correlation-driven cross-attention. We will briefly introduce these components in the remainder of this section. 

\subsection{Multi-Modal Physiological Signals Fusion}
Taking the DMER dataset as an example, after data pre-processing, we get a sampled input $x$ composed of four modalities, i.e., $x^{e} \in \mathbb{R}^{C_{e} \times d_{e}}$ for EEG, $x^{g} \in \mathbb{R}^{C_{g} \times d_{g}}$ for GSR, $x^{p} \in \mathbb{R}^{C_{p} \times d_{p}}$ for PPG, and $x^{v} \in \mathbb{R}^{C_{v} \times d_{v}}$ for video, where $C_{e}$, $C_{g}$, $C_{p}$, and $C_{v}$ denote the number of signal channels, and $d_{e}$, $d_{g}$, $d_{p}$, and $d_{v}$ are the extracted feature dimensions. As the heterogeneity within physiological signals (i.e., EEG, GSR, PPG, ECG, EDA, EMG) may be easier to handle than that of physiological and behavioral signals, we adopt a multi-head cross-attention mechanism to effectively fuse the physiological signals. Specifically, we first project EEG, GSR, and PPG signals into the same feature dimension. Then, we generate Queries on $x^e \in \mathbb{R}^{C \times d}$ and Keys, Values on $x^g \in \mathbb{R}^{C \times d}$ and $x^p \in \mathbb{R}^{C \times d}$, which can be formulated as follows:
\begin{equation}
\begin{gathered}
Q^{e}=x^{e} \mathbf{W}_Q^{e}, K^m=x^m \mathbf{W}_K^m, V^m=x^m \mathbf{W}_V^m
\end{gathered}
\end{equation}
where $\mathbf{W}_Q^{e}$, $\mathbf{W}_K^m$, $\mathbf{W}_V^m$ are learnable matrix, $m=\{g, p\}$ denotes the specific modality. Then, the cross-modal representation can be obtained by the self-attention mechanism:
\begin{equation}
\begin{aligned}
& x^{e \rightarrow g}=\operatorname{softmax}\left(\frac{Q^{e}\left(K^{g}\right)^T}{\sqrt{d}}\right) V^{g} \\
& x^{e \rightarrow p}=\operatorname{softmax}\left(\frac{Q^{e}\left(K^{p}\right)^T}{\sqrt{d}}\right) V^{p}
\end{aligned}
\end{equation}
where $d$ denotes a normalization parameter equal to the dimension of $K^m$. To learn multiple patterns of the association within physiological signals, multi-head attention mechanism and residual connection are employed:
\begin{equation}
x_{M H A}^{e \rightarrow m}=\left(x_{h_1}^{e \rightarrow m}\left\|x_{h_2}^{e \rightarrow m} \cdots\right\| x_{h_k}^{e \rightarrow m}\right) \mathbf{W}_{M H A}+L N\left(x^m\right)
\end{equation}
where $x_{M H A}^{e \rightarrow m} \in \mathbb{R}^{C \times \frac{d}{h}}$ is the output of multi-head cross attention block, $h$ is the number of heads, $\mathbf{W}_{MHA}$ denotes the transformation matrix, and $||$ represents the concatenation operation. Then, the concatenation of two cross-modal representations is adopted as the final physiological representation:
\begin{equation}
x^{p h y}=\left(x_{M H A}^{e \rightarrow g} \| x_{M H A}^{e \rightarrow p}\right)
\end{equation}

Using EEG as the query guides the cross-attention mechanism to focus on the emotional information at the neural level, effectively capturing the relationship between EEG and other physiological signals.
\subsection{Heterogeneity Mining with Optimal Transport}
Physiological signals, i.e., EEG, GSR, ECG, and PPG, provide subconscious emotional responses, while behavioral data reflects discrete and context-specific insights that are influenced by conscious control and environmental interactions. It is worth noting that different data characteristics lead to significant cross-modal heterogeneity, which heightens the difficulty of multi-modal fusion. Optimal transport (OT) \cite{perrot2016mapping,cao2022otkge,song2024multimodal} learns the optimal matching flow between two probability distributions, which allows for the effective alignment of heterogeneous multi-modal data distributions, facilitating the extraction of coherent patterns from physiological and behavioral signals. Formally, for the previous extracted physiological representations and projected behavioral representations, $x^{Phy} \in \mathbb{R}^{2C \times d}$ and $x^v \in \mathbb{R}^{2C \times d}$, an optimal transport from the $x^{Phy}$ to $x^{v}$ can be defined by the discrete Kantorovich formulation~\cite{kantorovich2006translocation} to search the overall optimal matching flow:
\begin{equation}
\mathcal{WD}\left(x^{P h y}, x^{v}\right)=\min _{f \in \Pi(u, \mathrm{v})}<T, \text { Cost }>_F
\end{equation}
where $\mathcal{WD}(,)$ denotes the wasserstein distance, $\Pi\left(u, v\right)=\{T \in \mathbb{R}_{+}^{2 C \times 2 C} \mid T 1_{2 C}=u, T^T 1_{2 C}=v\}$, $1_{2 C}$ represents a $2C$-dimensional all-one vector, $<,>_F$ refers to the Frobenius dot product, and $\text{Cost} \geq 0 \in \mathbb{R}^{2 C \times 2 C}$ is a cost matrix calculated by the cosine distance between $x^{Phy}$ and $x^{v}$. The matrix $T \in \mathbb{R}^{2C \times 2C}$ is denoted as the optimal transport matching flow. Then, we apply the Pytorch framework-based batch-wise implementation of the Sinkhorn algorithm proposed in~\cite{chen2020graph} to solve $\mathcal{WD}$. In this way, the obtained optimal transport matching flow $T$ can be used as a “cross-modal correlation” to enhance the multi-modal fusion. As such, the final multi-modal representation is adopted as follows:
\begin{equation}
x^m=(\text{Transformer}_l(x^{Phy}\otimes T)||\text{Transformer}_l(x^{v}))
\end{equation}
where $\text{Transformer}_l$ denotes a $l$-layer transformer encoder, $\otimes$ is the matrix multiplication, $||$ represents the concatenation operation. The optimal transport (OT) module bridges modality heterogeneity by aligning physiological and behavioral feature spaces through optimal matching flows, thereby mitigating distribution discrepancies.

\begin{figure}[]
\centering
\includegraphics[width=\linewidth]{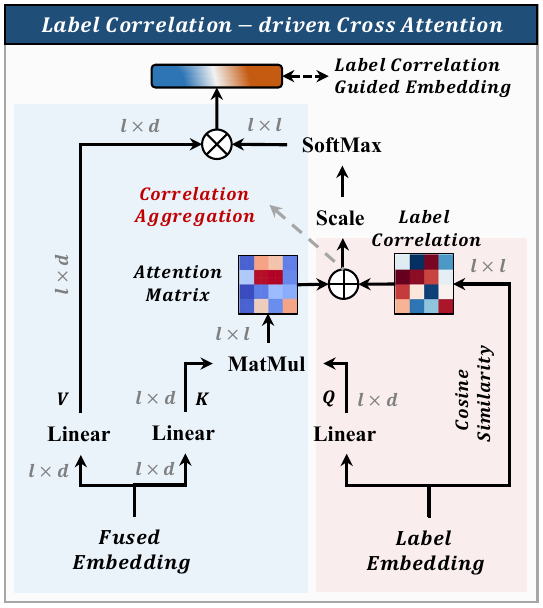}
\caption{\textbf{Illustration of the label correlation-driven cross-attention mechanism (LCDCA).}} \label{fig3}
\end{figure}

\subsection{Label Correlation-driven Cross-Attention}
In the emotion distribution learning scenario, correlation across different labels plays a vital role in the prediction of emotional distributions. Therefore, we design a label correlation-driven cross-attention mechanism (LCDCA), shown in Figure \ref{fig3}, to effectively enhance the prediction performance of the model. Specifically, we introduce a learnable label embedding $x^L \in \mathbb{R}^{l \times d}$, where $l$ is the number of emotion classes. To learn the label correlation, we first generate learnable label correlation $M^L \in \mathbb{R}^{l \times l}$ and ground-truth label correlation $M^{gt} \in \mathbb{R}^{l \times l}$ through the cosine similarity, whose $(i,j)$-th element is given as:
\begin{equation}
M_{i,j}^L=\frac{x_i^L\cdot x_j^L}{||x_i^L||\cdot||x_j^L||},M_{i,j}^{gt}=\frac{L_i\cdot L_j}{||L_i||\cdot||L_j||}
\label{eq7}
\end{equation}
where $L \in \mathbb{R}^{l \times 1}$ denotes the ground truth label distribution. Next, we adopt l-2 loss to constrain the learning of label correlation, and the correlation-constrained (CC) loss can be formulated as:
\begin{equation}
\mathcal{L}_{CC}=\|M^L-M^{gt}\|_2^2
\end{equation}

Then, we generate $Q^L$ on $x^L$, $K^m$ and $V^m$ on the projected multi-modal features $x^m$ through linear transformations. The label correlation-guided representation can be obtained by:
\begin{equation}
x^o=softmax(\frac{Q^L(K^m)^T+M^L}{\sqrt{d}})V^m
\end{equation}

With the LCDCA, the proposed model can effectively utilize the label correlation for emotion distribution learning. This design allows the model to dynamically aggregate global semantic relationships between features and labels, while simultaneously incorporating sample-specific correlations through attention matrix refinement. Finally, we adopt a prediction head consisting of a three-layer MLP and a softmax layer to generate the predicted emotion distribution $\hat{L}=(d_{1},d_{2},d_{3},\cdots,d_{l})$, where $l$ is the number of emotion classes. The Kullback-Leibler divergence (KLD) is adopted as the loss function to measure the distance between the predicted emotion distribution and the ground truth emotion distribution. Therefore, the overall training of our method is:
\begin{equation}
\mathcal{L}_{Overall}=\text{KLD}(\hat{L},L)+\mathcal{L}_{CC} 
\end{equation}
where $\text{KLD}(,)$ denotes the Kullback-Leibler divergence loss, $L$ is the ground truth emotion label distribution, and $\mathcal{L}_{CC}$ is the correlation constrained loss.

\begin{figure*}[!t]
\centering
\includegraphics[width=6.8in]{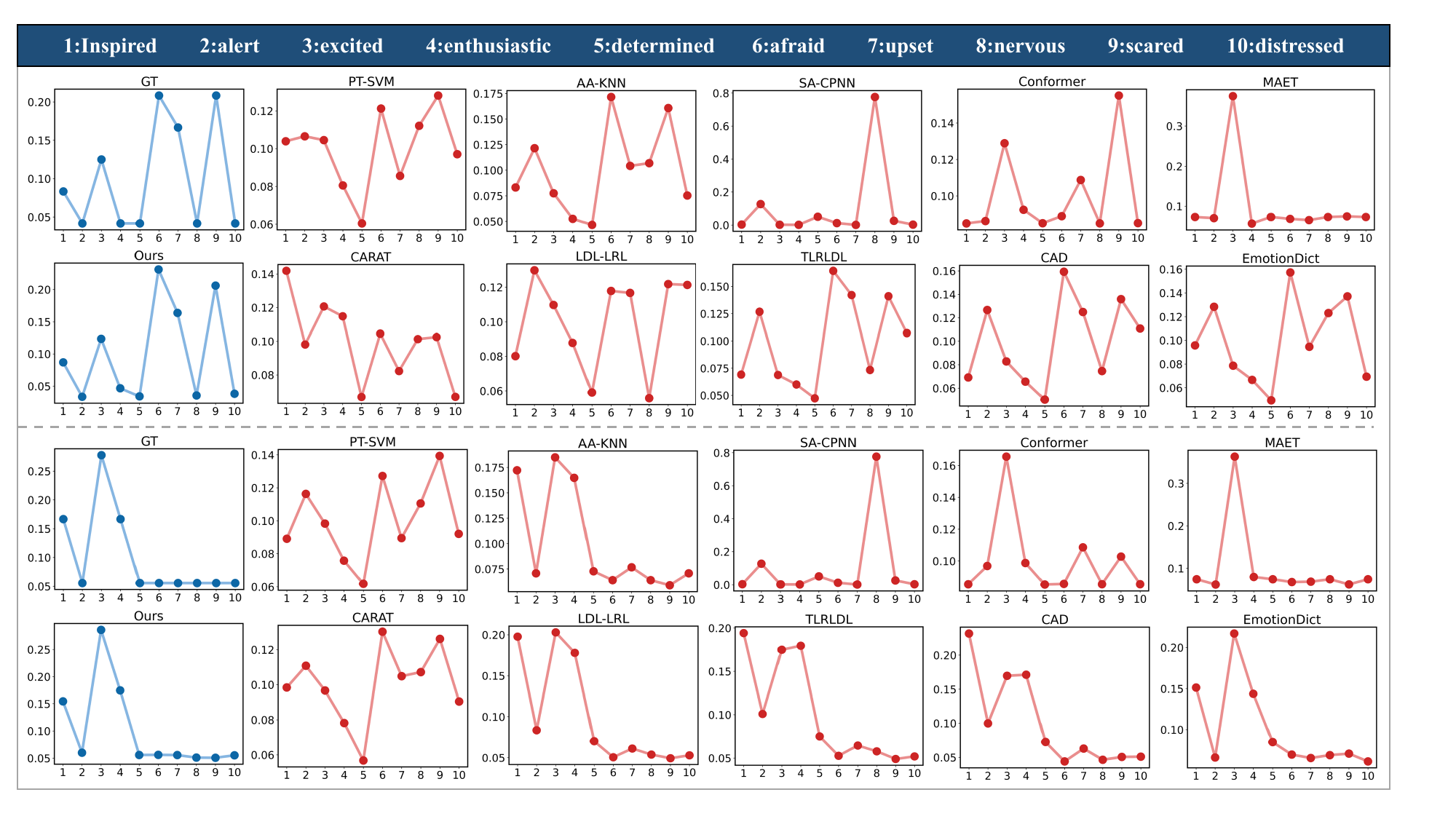}
\caption{\textbf{Predicted emotion distributions of our model and comparison methods.} We show two panels of two test samples of Subject No.45. GT indicates the ground truth distribution. The numbers 1 to 10 correspond to emotions inspired, alert, excited, enthusiastic, determined, afraid, upset, nervous, scared, and distressed.} \label{fig4}
\end{figure*}

\begin{table*}[h]
\caption{\textbf{Subject-dependent comparison of experimental results of our method and 10 comparison methods on six measures on the DMER and WESAD datasets.} ↓ indicates “the smaller the better”, and ↑ indicates “the larger the better”. The parentheses show the corresponding ranks on each evaluation metric and the \textit{Average Ranks}. The boldfaced scores denote the best performances and the underlined scores are the second runners. The results of the proposed HeLo are presented in a blue background.}
\resizebox{\linewidth}{!}{
\begin{tabular}{ccccccccccccc}
\toprule
\multirow{2}{*}{Dataset} & \multirow{2}{*}{Measure} & TKDE16~\cite{geng2016label}     & TKDE16~\cite{geng2016label}      & TPAMI13~\cite{geng2013facial}     & TNSRE22~\cite{song2022eeg}    & ACMMM23~\cite{jiang2023multimodal}            & AAAI24~\cite{peng2024carat}      & TKDE23~\cite{jia2023label}    & IJCAI24~\cite{ijcai2024p478}     & ICCV23~\cite{wen2023ordinal}      & TAFFC23~\cite{liu2023emotion}     & \multirow{2}{*}{\textbf{Ours}} \\ \cmidrule{3-12}
                         &                          & PT-SVM     & AA-KNN      & SA-CPNN     & Conformer  & MAET                & CARAT    & LDL-LRR   & TLRLDL      & CAD         & EmotionDict &                                \\ \midrule
\multirow{7}{*}{DMER}    & Chebyshev (↓)            & 0.0851 (7) & 0.0949 (10) & 0.0917 (8)  & 0.0661 (3) & \underline{0.0522 (2)}          & 0.0752 (4)  & 0.0936 (9) & 0.0847 (6)  & 0.1017 (11) & 0.0753 (5)  & \cellcolor{lightblue}\textbf{0.0446 (1)}            \\
                         & Clark (↓)                & 0.5787 (6) & 0.6649 (10) & 0.6333 (8)  & 0.4743 (3) & \underline{0.3762 (2)}          & 0.5951 (7)  & 0.6529 (9) & 0.5380 (4)  & 0.7517 (11) & 0.5463 (5)  & \cellcolor{lightblue}\textbf{0.3256 (1)}            \\
                         & Canberra (↓)             & 1.5294 (7) & 1.2357 (3)  & 1.8092 (9)  & 1.2581 (5) & \underline{0.9875 (2)}          & 1.5965 (8) & 1.8224 (10) & 1.3271 (5)  & 1.9415 (11) & 1.4771 (6)  & \cellcolor{lightblue}\textbf{0.8664 (1)}            \\
                         & KL (↓)                   & 0.0837 (6) & 0.1409 (10) & 0.1048 (8)  & 0.0593 (3) & \textbf{0.0288 (1)} & 0.1137 (9)  & 0.0644 (4) & 0.0925 (7)  & 0.1460 (11) & 0.0766 (5)  & \cellcolor{lightblue}\underline{0.0323 (2)}                     \\
                         & Cosine (↑)               & 0.9237 (8) & 0.9315 (6)  & 0.9058 (9)  & 0.9464 (3) & \underline{0.9688 (2)}          & 0.9332 (5) & 0.9016 (10) & 0.9339 (4)  & 0.8923 (11) & 0.9313 (7)  & \cellcolor{lightblue}\textbf{0.9714 (1)}            \\
                         & Intersection   (↑)       & 0.8417 (8) & 0.8425 (7)  & 0.8150 (9)  & 0.8716 (3) & \underline{0.9113 (2)}          & 0.8474 (6) & 0.8136 (10) & 0.8537 (5)  & 0.8077 (11) & 0.8492 (5)  & \cellcolor{lightblue}\textbf{0.9128 (1)}            \\ \cmidrule{2-13} 
                         & Average Rank (↓)       & 7 (7)      & 7.66 (8)    & 8.5 (9)     & 3.33 (3)   & \underline{1.83 (2)}            & 6.16 (6)    & 9.5 (10)   & 4.66 (4)    & 11 (11)     & 5.5 (5)     & \cellcolor{lightblue}\textbf{1.16 (1)}              \\ \midrule
\multirow{7}{*}{WESAD}   & Chebyshev (↓)            & 0.0515 (8) & 0.0346 (4)  & 0.0998 (11) & 0.0209 (3) & 0.0347 (5)          & 0.0354 (6)  & 0.0357 (7) & 0.0622 (10) & 0.0569 (9)  & \underline{0.0090 (2)}  & \cellcolor{lightblue}\textbf{0.0073 (1)}            \\
                         & Clark (↓)                & 0.3940 (8) & 0.2779 (4)  & 0.7285 (11) & 0.1838 (3) & 0.3503 (7)          & 0.2812 (5)  & 0.2822 (6) & 0.5381 (10) & 0.4618 (9)  & \underline{0.0733 (2)}  & \cellcolor{lightblue}\textbf{0.0653 (1)}            \\
                         & Canberra (↓)             & 1.0249 (8) & 0.7381 (4)  & 2.0746 (11) & 0.4810 (3) & 0.9100 (7)          & 0.7430 (5)  & 0.7466 (6) & 1.3586 (10) & 1.2199 (9)  & \underline{0.1876 (2)}  & \cellcolor{lightblue}\textbf{0.1614 (1)}            \\
                         & KL (↓)                   & 0.0320 (8) & 0.0243 (5)  & 0.1226 (10) & 0.0110 (3) & 0.0251 (6)          & 0.0210 (4)  & 0.0273 (7) & 0.2051 (11) & 0.0440 (9)  & 0.0017 (2)  & \cellcolor{lightblue}\textbf{0.0010 (1)}            \\
                         & Cosine (↑)               & 0.9728 (8) & 0.9793 (7)  & 0.8918 (11) & 0.9909 (3) & 0.9867 (4)          & 0.9824 (5)  & 0.9821 (6) & 0.9541 (10) & 0.9624 (9)  & \underline{0.9985 (2)}  & \cellcolor{lightblue}\textbf{0.9992 (1)}            \\
                         & Intersection   (↑)       & 0.9024 (8) & 0.9297 (4)  & 0.7896 (11) & 0.9550 (3) & 0.9213 (7)          & 0.9296 (5)  & 0.9292 (6) & 0.8746 (10) & 0.8812 (9)  & \underline{0.9820 (2)}  & \cellcolor{lightblue}\textbf{0.9905 (1)}            \\ \cmidrule{2-13} 
                         & Average Rank (↓)       & 8 (8)      & 5.33 (5)    & 10.83 (11)  & 3 (3)      & 6 (6)               & 5 (4)    & 6.33 (7)      & 10.16 (10)  & 9 (9)       & \underline{2.5 (2)}     & \cellcolor{lightblue}\textbf{1 (1)}                 \\ \bottomrule
\end{tabular}}
\label{tab:tab1}
\end{table*}

\begin{table*}[h]
\caption{\textbf{Subject-independent comparison of experimental results of our method and 10 comparison methods on six measures on the DMER and WESAD datasets.} ↓ indicates “the smaller the better”, and ↑ indicates “the larger the better”. The parentheses show the corresponding ranks on each evaluation metric and the \textit{Average Ranks}. The boldfaced scores denote the best performances and the underlined scores are the second runners. The results of the proposed HeLo are presented in a gray background.}
\resizebox{\linewidth}{!}{
\begin{tabular}{ccccccccccccc}
\toprule
\multirow{2}{*}{Dataset} & \multirow{2}{*}{Measure} & TKDE16~\cite{geng2016label}     & TKDE16~\cite{geng2016label}      & TPAMI13~\cite{geng2013facial}     & TNSRE22~\cite{song2022eeg}    & ACMMM23~\cite{jiang2023multimodal}            & AAAI24~\cite{peng2024carat}      & TKDE23~\cite{jia2023label}    & IJCAI24~\cite{ijcai2024p478}     & ICCV23~\cite{wen2023ordinal}      & TAFFC23~\cite{liu2023emotion}            & \multirow{2}{*}{\textbf{Ours}} \\ \cmidrule{3-12}
                         &                          & PT-SVM      & AA-KNN      & SA-CPNN     & Conformer  & MAET        & CARAT    & LDL-LRR     & TLRLDL      & CAD        & EmotionDict         &                                \\ \midrule
\multirow{7}{*}{DMER}    & Chebyshev (↓)            & 0.1462 (11) & 0.1070 (10) & 0.0928 (3)  & 0.0933 (4) & 0.0939 (5)  & 0.1023 (9) & 0.1012 (8)  & 0.0964 (6)  & \underline{0.0925 (2)} & 0.0945 (7)          & \cellcolor{lightblue}\textbf{0.0882 (1)}            \\
                         & Clark (↓)                & 1.2877 (11) & 0.6860 (9)  & 0.6440 (3)  & 0.6515 (4) & 0.6605 (5)  & 0.6909 (10) & 0.6856 (8) & 0.6702 (6)  & \underline{0.6379 (2)} & 0.6676 (7)          & \cellcolor{lightblue}\textbf{0.6289 (1)}            \\
                         & Canberra (↓)             & 2.0459 (11) & 1.8308 (7)  & 1.8542 (8)  & \underline{1.7689 (2)} & 1.7863 (3)  & 1.9161 (10) & 1.9071 (9) & 1.8244 (6)  & 1.8104 (5) & 1.8092 (4)          & \cellcolor{lightblue}\textbf{1.7603 (1)}            \\
                         & KL (↓)                   & 0.5528 (11) & 0.1227 (9)  & 0.1089 (5)  & \underline{0.1061 (2)} & 0.1088 (4)  & 0.1245 (10) & 0.1222 (8) & 0.1142 (7)  & 0.1076 (3) & 0.1104 (6)          & \cellcolor{lightblue}\textbf{0.1027 (1)}            \\
                         & Cosine (↑)               & 0.8295 (11) & 0.8843 (10) & 0.9022 (4)  & \underline{0.9038 (2)} & 0.9016 (5)  & 0.8899 (9) & 0.8917 (8)  & 0.8964 (7)  & 0.9036 (3) & 0.9001 (6)          & \cellcolor{lightblue}\textbf{0.9148 (1)}            \\
                         & Intersection (↑)       & 0.7762 (11) & 0.8050 (8)  & 0.8102 (6)  & \underline{0.8161 (2)} & 0.8147 (3)  & 0.8026 (10) & 0.8037 (9) & 0.8087 (7)  & 0.8122 (4) & 0.8119 (5)          & \cellcolor{lightblue}\textbf{0.8194 (1)}            \\ \cmidrule{2-13} 
                         & Average Rank   (↓)       & 11 (11)     & 8.83 (9)    & 4.83 (5)    & \underline{2.66 (2)}   & 4.16 (4)    & 9.66 (10)   & 8.33 (8)   & 6.5 (7)     & 3.16 (3)   & 5.83 (6)            & \cellcolor{lightblue}\textbf{1 (1)}                 \\ \midrule
\multirow{7}{*}{WESAD}   & Chebyshev (↓)            & 0.0471 (7)  & 0.0500 (9)  & 0.0894 (11) & 0.0436 (3) & 0.0516 (10) & 0.0454 (5) & 0.0466 (6)  & 0.0441 (4)  & 0.0481 (8) & \underline{0.0421 (2)}          & \cellcolor{lightblue}\textbf{0.0403 (1)}            \\
                         & Clark (↓)                & 0.4506 (8)  & 0.4516 (10) & 0.6394 (11) & 0.4048 (3) & 0.4514 (9)  & 0.4270 (5) & 0.4471 (6)  & 0.4145 (4)  & 0.4488 (7) & \underline{0.3631 (2)}          & \cellcolor{lightblue}\textbf{0.3455 (1)}            \\
                         & Canberra (↓)             & 1.1928 (10) & 1.1762 (7)  & 1.7384 (11) & 1.0806 (4) & 1.1666 (6)  & 1.1874 (9) & 1.1817 (8)  & 1.0746 (3)  & 1.1647 (5) & \underline{0.9530 (2)}          & \cellcolor{lightblue}\textbf{0.9329 (1)}            \\
                         & KL (↓)                   & 0.0347 (5)  & 0.0423 (9)  & 0.0944 (11) & 0.0369 (7) & 0.0445 (10) & 0.0345 (4) & 0.0343 (3)  & 0.0372 (8)  & 0.0366 (6) & \textbf{0.0282 (1)} & \cellcolor{lightblue}\underline{0.0283 (2)}                     \\
                         & Cosine (↑)               & 0.9727 (5)  & 0.9663 (10) & 0.9154 (11) & 0.9699 (7) & 0.9678 (9)  & 0.9731 (3) & 0.9731 (3)  & 0.9695 (8)  & 0.9711 (6) & \underline{0.9773 (2)}          & \cellcolor{lightblue}\textbf{0.9790 (1)}            \\
                         & Intersection   (↑)       & 0.8935 (7)  & 0.8933 (9)  & 0.8254 (11) & 0.9015 (3) & 0.8967 (4)  & 0.8944 (6) & 0.8945 (5)  & 0.8917 (10) & 0.8935 (7) & \underline{0.9126 (2)}          & \cellcolor{lightblue}\textbf{0.9154 (1)}            \\ \cmidrule{2-13} 
                         & Average Rank (↓)       & 7 (8)       & 9 (10)      & 11 (11)     & 4.5 (3)    & 8 (9)       & 5.33 (5)   & 5.16 (4)    & 6.16 (6)    & 6.5 (7)    & \underline{1.83 (2)}            & \cellcolor{lightblue}\textbf{1.16 (1)}              \\ \bottomrule
\end{tabular}}

\label{tab:tab2}
\end{table*}

\section{Experiment}

\subsection{Multi-Modal Emotion Datasets}
We evaluate our proposed method on two publicly available datasets: \textbf{DMER}~\cite{yang2024multimodal} and \textbf{WESAD}~\cite{schmidt2018introducing} datasets. \textbf{DMER} dataset contains four modalities including EEG, galvanic skin response (GSR), photoplethysmogram (PPG), and facial videos collected from 80 participants. Each participant was asked to watch 32 video clips. In each trial, the subject first watched one video clip, and then completed the 10-item short positive and negative affect schedules (PANAS~\cite{watson1994panas}). The score for each basic emotion ranged from 1 (none) to 5 (strong) and was then transformed into emotion distributions. The data of 73 subjects were used for further experiments in this paper. \textbf{WESAD} contains ECG, EMG, electrodermal activity (EDA), and 3-axis accelerometer (ACC) from 17 participants. The data were collected at a sampling rate of 700Hz, and the goal was to study four different affective states namely neutral, stressed, amused, and meditated. Upon completion of each trial, the ground truth labels for the affect states were collected using the PANAS~\cite{watson1994panas}. 14 participants were utilized in our experiments. Notably, ECG, EMG, and EDA are the "Physiological data", and ACC denotes the "Behavioral Data" in our proposed method. Details of the data preprocessing can be referred to the \textbf{\textit{supplementary materials}}.

\subsection{Evaluation Metrics}
For the evaluation of our proposed model, following the previous emotion distribution learning studies \cite{shu2022emotion,liu2023emotion,geng2016label}, we adopt six distribution-based measurements, including four distance metrics (Chebyshev (↓), Clark (↓), Canberra (↓), and Kullback-Leibler (KL) (↓)) and two similarity measurements (Cosine (↑) and Intersection (↑)). ↓ indicates the metric is “the smaller the better”, and ↑ denotes “the larger the better”. Moreover, we reported the Average Rank denoting the mean rank of the six metrics following \cite{shu2022emotion,liu2023emotion}. 

\subsection{Implementation Details}
In our experiments, following previous works \cite{shu2022emotion,liu2023emotion}, both subject-dependent and subject-independent settings are adopted to verify the method on the two datasets. In subject-dependent recognition, the data of each subject are split into the training set and testing set in a ratio of 8:2 randomly, and the average performance across all subjects is taken as the final result. For subject-independent recognition, we utilize the leave-one-subject-out cross-validation for the evaluation, and the average performance across all iterations is reported as the final result. Our proposed framework is trained on a 12GB NVIDIA GeForce RTX3080 GPU and developed under PyTorch. The training adopts Adam optimizer with a learning rate of $10^{-3}$. The batch size is set to 128, the number of epochs for training is set to 300, and the number of attention heads is 4. The embedding size in the attention mechanism and feed-forward network is set to 128, and 64, respectively, and we set the depth of transformer blocks to 1. Our code is released at: https://github.com/kaio-99/HeLo.

\subsection{Comparison with State-of-the-art Methods}

To evaluate the effectiveness of our proposed method, we compare our model with several state-of-the-art models on both the DMER and AMIGOS datasets. These methods including label distribution learning methods (PT-SVM~\cite{geng2016label}, AA-KNN~\cite{geng2016label}, SA-CPNN~\cite{geng2013facial}), EEG-based classification method (Conformer~\cite{song2022eeg}), multi-modal emotion recognition method (MAET~\cite{jiang2023multimodal}), multi-modal multi-label emotion recognition method (CARAT~\cite{peng2024carat}), recently proposed state-of-the-art single-modal label distribution learning methods (LDL-LRR~\cite{jia2023label}, TLRLDL~\cite{ijcai2024p478}, CAD~\cite{wen2023ordinal}), and multi-modal emotion distribution learning method (EmotionDict~\cite{liu2023emotion}). 

The results of the subject-dependent recognition on the DMER and WESAD datasets are summarized in Table \ref{tab:tab1}. As can be seen from the table, except for the KL metric on the DMER dataset, our proposed method outperforms all the other state-of-the-art methods over the six metrics on both datasets. Benefiting from deep learning's ability to extract high-dimensional features, the deep learning-based methods achieve relatively better performance than the traditional LDL methods (PT-SVM, AA-KNN, and SA-CPNN). Besides, visualization examples of the predicted emotion distributions using all the methods and the ground truths on the DMER dataset are shown in Figure \ref{fig4}. It can be seen that some methods, i.e., PT-SVM, SA-CPNN, Conformer, and MAET, are more likely to output similar distributions for different signal samples, leading to sub-optimal emotion distribution learning results. The qualitative results show that our method can predict emotion distributions most similar to the ground truths. The superiority of our method lies in that our model not only effectively utilizes the fusion of heterogeneous multi-modal data, but also leverages the learnable label correlation for the prediction of emotion distribution.

\begin{table}[]
\caption{\textbf{Ablation study in modalities on the DMER.}}
\resizebox{\linewidth}{!}{
\begin{tabular}{ccccccc}
\toprule
\multirow{2}{*}{Settings}                                                                   & \multirow{2}{*}{Measure} & \multirow{2}{*}{w/o EEG} & \multirow{2}{*}{w/o GSR} & \multirow{2}{*}{w/o PPG} & \multirow{2}{*}{w/o Video} & \multirow{2}{*}{\textbf{Ours}} \\
                                                                                            &                          &                          &                          &                          &                            &                                \\ \midrule
\multirow{6}{*}{\begin{tabular}[c]{@{}c@{}}Subject-Dependent\end{tabular}}    & Chebyshev (↓)            & 0.0529                   & 0.0574                   & 0.0487                   & 0.0552                     & \cellcolor{lightblue}\textbf{0.0446}                \\
                                                                                            & Clark (↓)                & 0.3372                   & 0.3550                   & 0.3671                   & 0.3380                     & \cellcolor{lightblue}\textbf{0.3256}                \\
                                                                                            & Canberra (↓)             & 0.9028                   & 0.8987                   & 0.8820                   & 0.9065                     & \cellcolor{lightblue}\textbf{0.8664}                \\
                                                                                            & KL (↓)                   & 0.0381                   & 0.0429                   & 0.0357                   & 0.0377                     & \cellcolor{lightblue}\textbf{0.0323}                \\
                                                                                            & Cosine (↑)               & 0.9677                   & 0.9591                   & 0.9692                   & 0.9628                     & \cellcolor{lightblue}\textbf{0.9714}                \\
                                                                                            & Intersection   (↑)       & 0.9102                   & 0.9007                   & 0.9086                   & 0.9033                     & \cellcolor{lightblue}\textbf{0.9128}                \\ \midrule
\multirow{6}{*}{\begin{tabular}[c]{@{}c@{}}Subject-Independent\end{tabular}}  & Chebyshev (↓)            & 0.0977                   & 0.0953                   & 0.0964                   & 0.0926                     & \cellcolor{lightblue}\textbf{0.0882}                \\
                                                                                            & Clark (↓)                & 0.6906                   & 0.6742                   & 0.6470                   & 0.6635                     & \cellcolor{lightblue}\textbf{0.6289}                \\
                                                                                            & Canberra (↓)             & 1.9261                   & 1.9077                   & 1.8074                   & 1.8250                     & \cellcolor{lightblue}\textbf{1.7603}                \\
                                                                                            & KL (↓)                   & 0.1175                   & 0.1244                   & 0.1085                   & 0.1134                     & \cellcolor{lightblue}\textbf{0.1027}                \\
                                                                                            & Cosine (↑)               & 0.9080                   & 0.9075                   & 0.9122                   & 0.9107                     & \cellcolor{lightblue}\textbf{0.9148}                \\
                                                                                            & Intersection   (↑)       & 0.8197                   & 0.8056                   & 0.8138                   & 0.8121                     & \cellcolor{lightblue}\textbf{0.8194}                    \\ \bottomrule
\end{tabular}}
\label{tab:tab3}
\end{table}

\begin{table}[]
\caption{\textbf{Ablation study in modalities on the WESAD.}}
\resizebox{\linewidth}{!}{
\begin{tabular}{ccccccc}
\toprule
\multirow{2}{*}{Settings}                                                                   & \multirow{2}{*}{Measure} & \multirow{2}{*}{w/o ECG} & \multirow{2}{*}{w/o EMG} & \multirow{2}{*}{w/o EDA} & \multirow{2}{*}{w/o ACC} & \multirow{2}{*}{\textbf{Ours}} \\
                                                                                            &                          &                          &                          &                          &                            &                                \\ \midrule
\multirow{6}{*}{\begin{tabular}[c]{@{}c@{}}Subject-Dependent\end{tabular}}   & Chebyshev (↓)            & 0.0088                   & 0.0074                   & 0.0079                   & 0.0082                     & \cellcolor{lightblue}\textbf{0.0073}                \\
                                                                                            & Clark (↓)                & 0.0744                   & 0.0722                   & 0.0762                   & 0.0695                     & \cellcolor{lightblue}\textbf{0.0653}                \\
                                                                                            & Canberra (↓)             & 0.1730                   & 0.1630                   & 0.1658                   & 0.1725                     & \cellcolor{lightblue}\textbf{0.1614}                \\
                                                                                            & KL (↓)                   & 0.0022                   & 0.0016                   & 0.0014                   & 0.0016                     & \cellcolor{lightblue}\textbf{0.0010}                \\
                                                                                            & Cosine (↑)               & 0.9915                   & 0.9960                   & 0.9955                   & 0.9971                     & \cellcolor{lightblue}\textbf{0.9992}                \\
                                                                                            & Intersection   (↑)       & 0.9873                   & 0.9899                   & 0.9892                   & 0.9885                     & \cellcolor{lightblue}\textbf{0.9905}                \\ \midrule
\multirow{6}{*}{\begin{tabular}[c]{@{}c@{}}Subject-Independent\end{tabular}} & Chebyshev (↓)            & 0.0502                   & 0.0419                   & 0.0429                   & 0.0474                     & \cellcolor{lightblue}\textbf{0.0403}                \\
                                                                                            & Clark (↓)                & 0.3877                   & 0.3597                   & 0.3721                   & 0.3842                     & \cellcolor{lightblue}\textbf{0.3455}                \\
                                                                                            & Canberra (↓)             & 1.0748                   & 0.9516                   & 0.9579                   & 1.0029                     & \cellcolor{lightblue}\textbf{0.9329}                \\
                                                                                            & KL (↓)                   & 0.0322                   & 0.0308                   & 0.0344                   & 0.0315                     & \cellcolor{lightblue}\textbf{0.0283}                \\
                                                                                            & Cosine (↑)               & 0.9688                   & 0.9722                   & 0.9641                   & 0.9658                     & \cellcolor{lightblue}\textbf{0.9790}                \\
                                                                                            & Intersection   (↑)       & 0.8870                   & 0.9095                   & 0.9017                   & 0.8991                     & \cellcolor{lightblue}\textbf{0.9154}                \\ \bottomrule
\end{tabular}}
\label{tab:tab4}
\end{table}

Table \ref{tab:tab2} illustrates the subject-independent evaluation of the comparison methods and our proposed model on the DMER datasets. Except for the KL metric on the WESAD dataset, our method achieves the best results across both datasets and shows the highest average rank. The subject-independent recognition is more challenging than the subject-dependent situation, showing the superiority of our proposed method in capturing the subject-invariant emotional features across different individuals.


\begin{figure}[htbp]
  \centering
  {\includegraphics[width=0.7\linewidth]{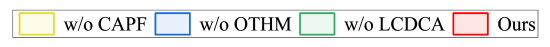}}\\
  \subfloat[DMER Subject-Dependent]
  {\includegraphics[width=0.47\linewidth]{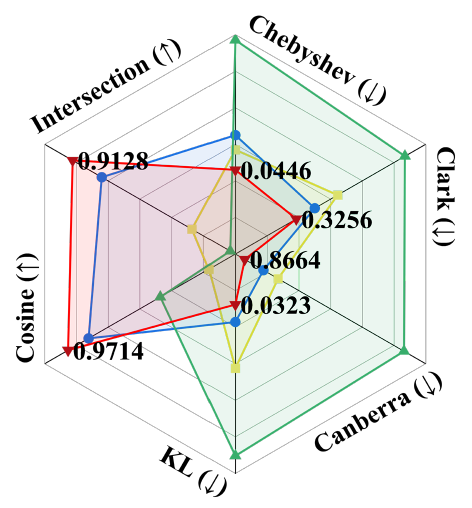}}\label{fig:5-1}
  \subfloat[DMER Subject-Indpendent]
  {\includegraphics[width=0.47\linewidth]{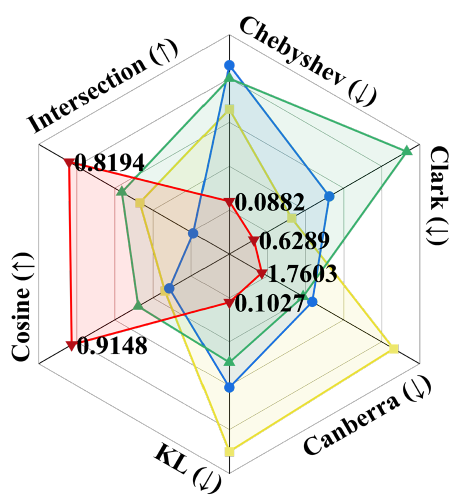}}\label{fig:5-2}\\
  \subfloat[WESAD Subject-Dependent]
  {\includegraphics[width=0.47\linewidth]{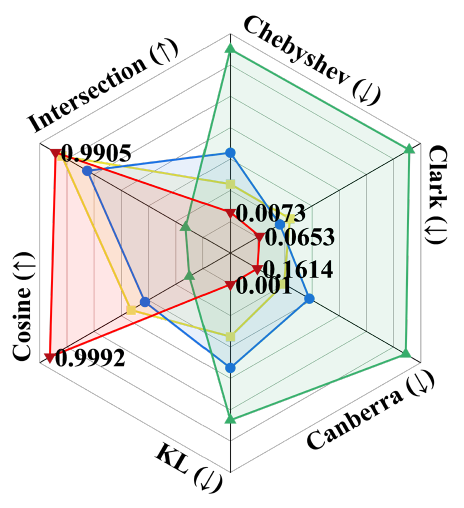}}\label{fig:5-3}
  \subfloat[WESAD Subject-Independent]
  {\includegraphics[width=0.47\linewidth]{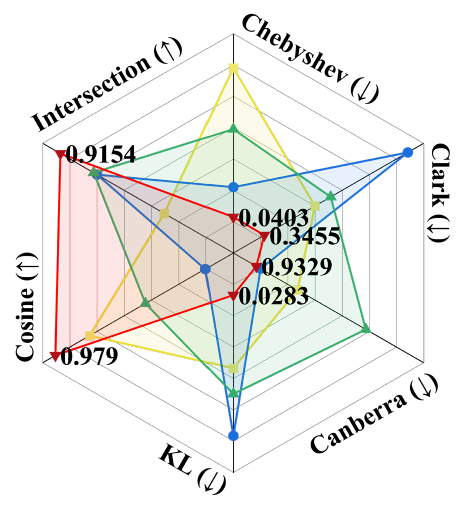}}\label{fig:5-4}
  \caption{\textbf{Ablation of network components in HeLo}.}
  \label{fig5}
\end{figure}

\subsection{Ablation Studies}
\textbf{Ablation on network components.} To verify the rationality of the modules of the proposed framework, we conduct ablation experiments on the DMER and WESAD datasets by removing each module of our method, i.e., cross-attention-based physiological signals fusion (CAPF), optimal transport-based heterogeneity mining (OTHM), and label-correlation-driven cross-attention (LCDCA), and the results are illustrated in Figure \ref{fig5}. The experimental results show that each module in the model has a positive contribution to the emotion distribution learning performance of the model, and the proposed method achieves the best performance through the effective integration of each module. CAPF effectively fuses the physiological signals to enhance the model’s performance. OTHM achieves better performance than concatenation, indicating its proficiency in capturing heterogeneous information across physiological and behavioral features. Furthermore, LCDCA greatly improves the model’s performance as it integrates the label correlations for the learning of emotion distribution.

\noindent \textbf{Ablation on the modalities.} We also assessed the impact of each modality by separately removing each modality from our full model, the results are shown in Table \ref{tab:tab3} and \ref{tab:tab4}. These experiments were conducted in both subject-dependent and subject-independent settings across both datasets. The results indicate that each modality contributes significantly to the overall performance of multi-modal emotion distribution learning, effectively integrating multiple modalities can enhance the emotion distribution performance. 

\begin{figure}[]
\centering
\includegraphics[width=3.3in]{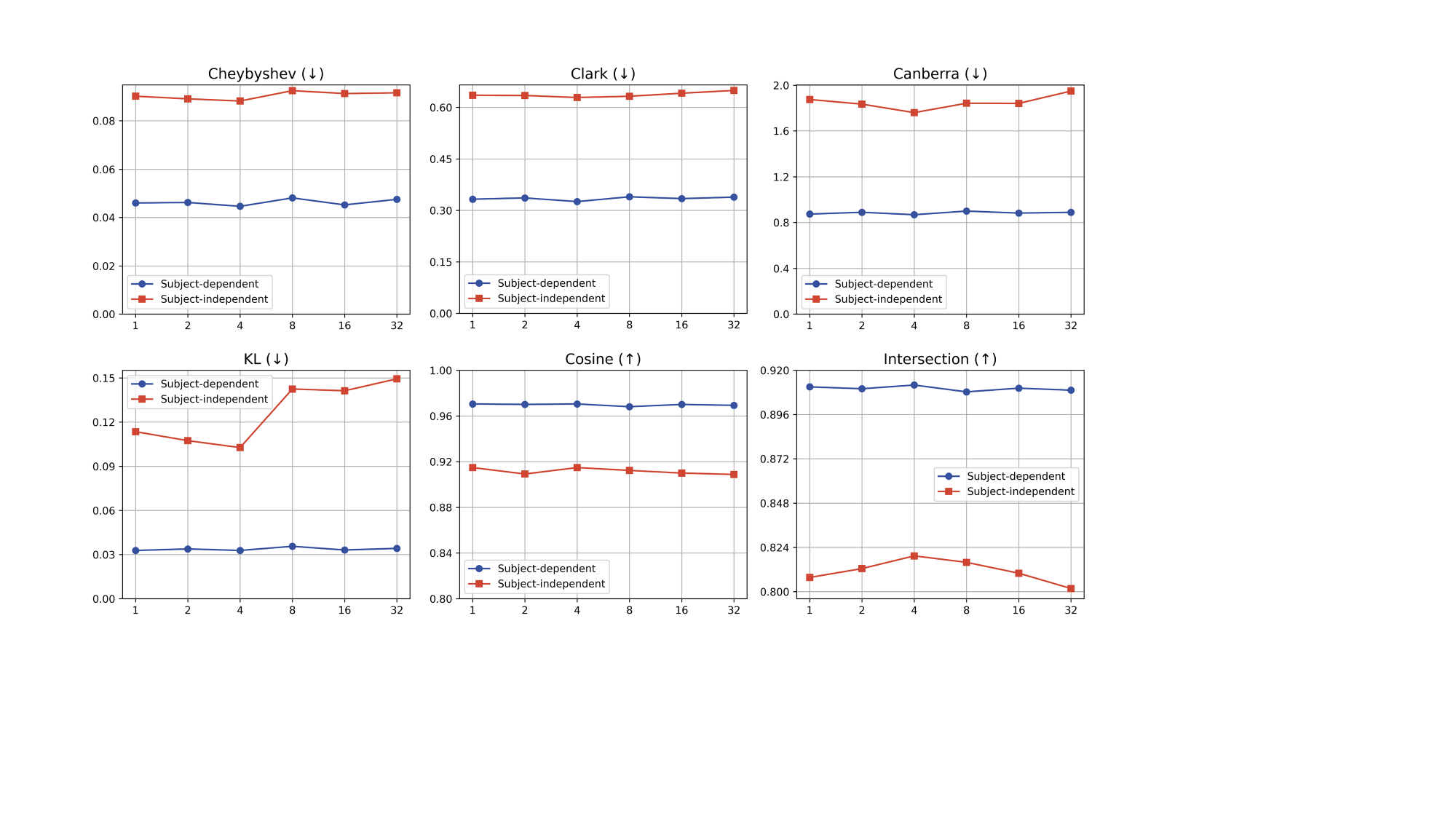}
\caption{\textbf{Effect of attention heads on the DMER}} \label{fig6}
\end{figure}

\begin{figure}[]
\centering
\includegraphics[width=3.3in]{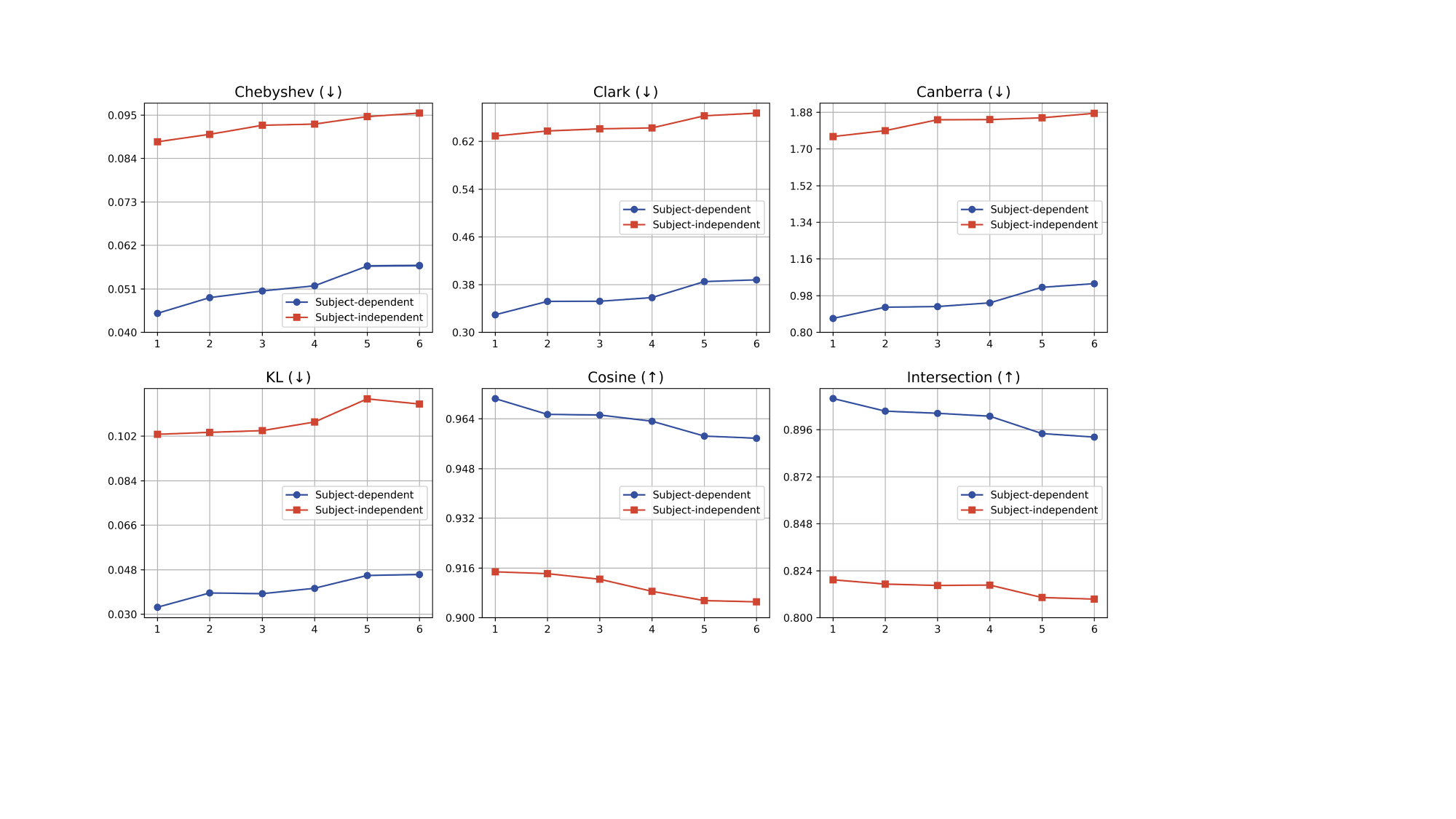}
\caption{Effect of transformer block depth on the DMER.} \label{fig7}
\end{figure}

\noindent \textbf{Ablation on the number of attention heads.} The number of attention heads is usually an essential parameter for models based on attention mechanisms. The result in Figure \ref{fig6} indicates that, the model's performance in both subject-dependent and subject-independent settings initially improves as the number of heads increases, and subsequently declines. The model reaches its optimal performance across all six evaluation metrics when the number of attention heads was 4. However, the overall impact of the number of attention heads on the model's performance is relatively limited.

\noindent \textbf{Ablation on the depth of transformer block.} The depth of end-to-end models is considered a crucial factor in determining their fitting ability. Figure \ref{fig7} illustrates that, the model achieved optimal classification performance when the transformer block depth was 1. As the depth increased, the classification accuracy showed a slight tendency to decrease and then leveled off. This is because a greater transformer block depth might cause overfitting.

\begin{figure}[h]
\centering
\includegraphics[width=3.3in]{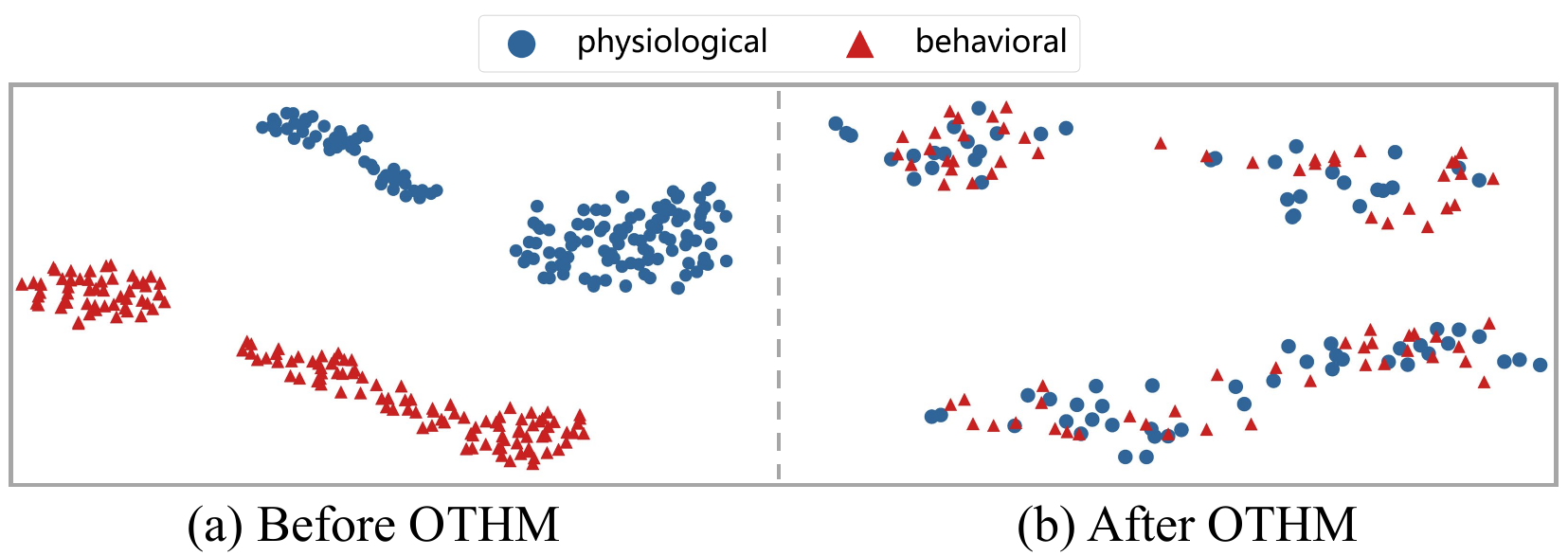}
\caption{\textbf{Visualization of the feature embedding.}} \label{fig8}
\end{figure}

\begin{figure}[]
\centering
\includegraphics[width=3.3in]{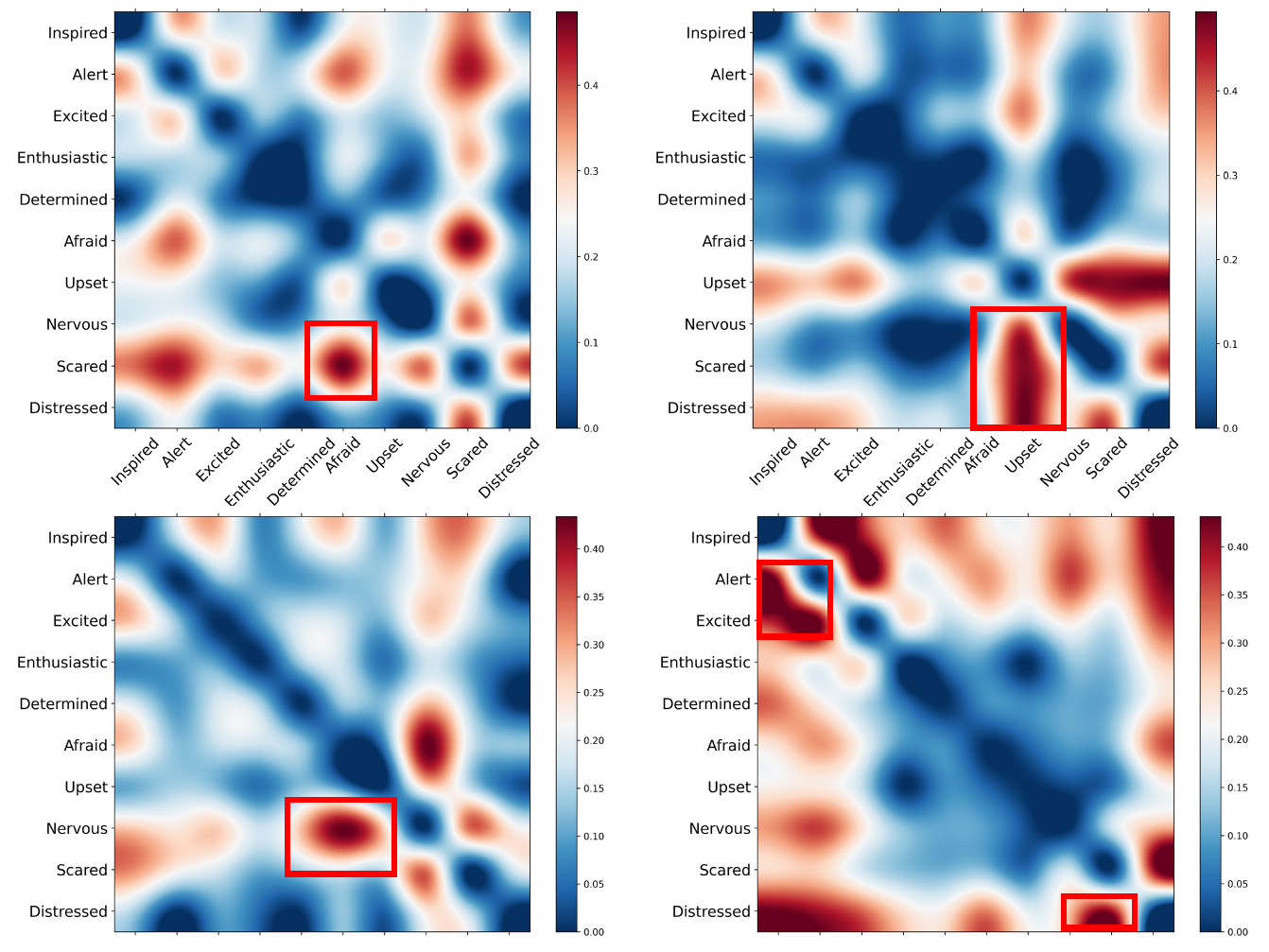}
\caption{\textbf{Label correlations visualization.} A higher red intensity value indicates a stronger correlation.} \label{fig9}
\end{figure}

\subsection{Visualization}
\textbf{Visualization of the cross-modal representations.} We visualize the multi-modal feature space in the testing set by utilizing t-SNE~\cite{van2008visualizing}. In Figure \ref{fig8} (a), before the OTHM module, there is a noticeable separation between physiological (blue) and behavioral (red) representations. This pronounced heterogeneity illustrates the challenges faced in multi-modal fusion due to distinct feature distributions, which can impede effective integration and analysis. With the help of OTHM, shown in Figure \ref{fig8} (b), the cross-modal representations exhibit a closer arrangement and overlap more significantly than before. This reduction in heterogeneity suggests that OTHM has effectively minimized the disparities between the physiological and behavioral features, facilitating a better-aligned feature space and a more coherent multi-modal fusion.

\noindent \textbf{Visualization of the learned label correlations.} To verify the effectiveness of the proposed LCDCA module, we visualize the learned label correlation $M^L$ in Figure \ref{eq7} to illustrate the interpretability, and the results on the testing set of DMER dataset are shown in Figure \ref{fig9}. As can be seen from Figure \ref{fig9}, the label correlations differ from different trials and different subjects, which jointly attend to rich semantic information from various perspectives. From the horizontal view, “afraid” is highly correlated with “nervous”, “scared”. Besides, “enthusiastic”, and “determined” are usually correlated with “excited”, “alert”. All of these conclusions are consistent with human intuition, so as to demonstrate the better emotion distribution learning performance of our proposed model.



\section{Conclusion}
In this paper, we introduce a novel multi-modal emotion distribution learning framework, HeLo, which effectively addresses the challenges of capturing complex, mixed emotional states from heterogeneous physiological and behavioral data. The CAPF module ensures the robust fusion of multi-modal physiological signals, and the OTHM module efficiently mines the heterogeneity between physiological and behavioral representations, bridging the gap between these diverse data sources and enhancing feature alignment. Moreover, the LCDCA fully exploits label correlations, guiding the learning of emotion distributions in a semantically meaningful way. Experimental results on two multi-modal emotion datasets demonstrate the efficacy of our proposed method.

\bibliographystyle{ACM-Reference-Format}
\bibliography{sample-base}


\begin{thebibliography}{52}


\ifx \showCODEN    \undefined \def \showCODEN     #1{\unskip}     \fi
\ifx \showISBNx    \undefined \def \showISBNx     #1{\unskip}     \fi
\ifx \showISBNxiii \undefined \def \showISBNxiii  #1{\unskip}     \fi
\ifx \showISSN     \undefined \def \showISSN      #1{\unskip}     \fi
\ifx \showLCCN     \undefined \def \showLCCN      #1{\unskip}     \fi
\ifx \shownote     \undefined \def \shownote      #1{#1}          \fi
\ifx \showarticletitle \undefined \def \showarticletitle #1{#1}   \fi
\ifx \showURL      \undefined \def \showURL       {\relax}        \fi
\providecommand\bibfield[2]{#2}
\providecommand\bibinfo[2]{#2}
\providecommand\natexlab[1]{#1}
\providecommand\showeprint[2][]{arXiv:#2}

\bibitem[Alarcao and Fonseca(2017)]%
        {alarcao2017emotions}
\bibfield{author}{\bibinfo{person}{Soraia~M Alarcao} {and}
  \bibinfo{person}{Manuel~J Fonseca}.} \bibinfo{year}{2017}\natexlab{}.
\newblock \showarticletitle{Emotions recognition using EEG signals: A survey}.
\newblock \bibinfo{journal}{\emph{IEEE Transactions on Affective Computing}}
  \bibinfo{volume}{10}, \bibinfo{number}{3} (\bibinfo{year}{2017}),
  \bibinfo{pages}{374--393}.
\newblock


\bibitem[Can et~al\mbox{.}(2023)]%
        {can2023approaches}
\bibfield{author}{\bibinfo{person}{Yekta~Said Can}, \bibinfo{person}{Bhargavi
  Mahesh}, {and} \bibinfo{person}{Elisabeth Andr{\'e}}.}
  \bibinfo{year}{2023}\natexlab{}.
\newblock \showarticletitle{Approaches, applications, and challenges in
  physiological emotion recognition—a tutorial overview}.
\newblock \bibinfo{journal}{\emph{Proc. IEEE}} (\bibinfo{year}{2023}).
\newblock


\bibitem[Cao et~al\mbox{.}(2022)]%
        {cao2022otkge}
\bibfield{author}{\bibinfo{person}{Zongsheng Cao}, \bibinfo{person}{Qianqian
  Xu}, \bibinfo{person}{Zhiyong Yang}, \bibinfo{person}{Yuan He},
  \bibinfo{person}{Xiaochun Cao}, {and} \bibinfo{person}{Qingming Huang}.}
  \bibinfo{year}{2022}\natexlab{}.
\newblock \showarticletitle{Otkge: Multi-modal knowledge graph embeddings via
  optimal transport}.
\newblock \bibinfo{journal}{\emph{Advances in Neural Information Processing
  Systems}}  \bibinfo{volume}{35} (\bibinfo{year}{2022}),
  \bibinfo{pages}{39090--39102}.
\newblock


\bibitem[Chen et~al\mbox{.}(2020)]%
        {chen2020graph}
\bibfield{author}{\bibinfo{person}{Liqun Chen}, \bibinfo{person}{Zhe Gan},
  \bibinfo{person}{Yu Cheng}, \bibinfo{person}{Linjie Li},
  \bibinfo{person}{Lawrence Carin}, {and} \bibinfo{person}{Jingjing Liu}.}
  \bibinfo{year}{2020}\natexlab{}.
\newblock \showarticletitle{Graph optimal transport for cross-domain
  alignment}. In \bibinfo{booktitle}{\emph{International Conference on Machine
  Learning}}. PMLR, \bibinfo{pages}{1542--1553}.
\newblock


\bibitem[Cheng and Liu(2008)]%
        {cheng2008emotion}
\bibfield{author}{\bibinfo{person}{Bo Cheng} {and} \bibinfo{person}{Guangyuan
  Liu}.} \bibinfo{year}{2008}\natexlab{}.
\newblock \showarticletitle{Emotion recognition from surface EMG signal using
  wavelet transform and neural network}. In \bibinfo{booktitle}{\emph{2008 2nd
  International Conference on Bioinformatics and Biomedical Engineering}}.
  IEEE, \bibinfo{pages}{1363--1366}.
\newblock


\bibitem[Gao et~al\mbox{.}(2017)]%
        {gao2017deep}
\bibfield{author}{\bibinfo{person}{Bin-Bin Gao}, \bibinfo{person}{Chao Xing},
  \bibinfo{person}{Chen-Wei Xie}, \bibinfo{person}{Jianxin Wu}, {and}
  \bibinfo{person}{Xin Geng}.} \bibinfo{year}{2017}\natexlab{}.
\newblock \showarticletitle{Deep label distribution learning with label
  ambiguity}.
\newblock \bibinfo{journal}{\emph{IEEE Transactions on Image Processing}}
  \bibinfo{volume}{26}, \bibinfo{number}{6} (\bibinfo{year}{2017}),
  \bibinfo{pages}{2825--2838}.
\newblock


\bibitem[Geng(2016)]%
        {geng2016label}
\bibfield{author}{\bibinfo{person}{Xin Geng}.} \bibinfo{year}{2016}\natexlab{}.
\newblock \showarticletitle{Label distribution learning}.
\newblock \bibinfo{journal}{\emph{IEEE Transactions on Knowledge and Data
  Engineering}} \bibinfo{volume}{28}, \bibinfo{number}{7}
  (\bibinfo{year}{2016}), \bibinfo{pages}{1734--1748}.
\newblock


\bibitem[Geng and Ji(2013)]%
        {geng2013label}
\bibfield{author}{\bibinfo{person}{Xin Geng} {and} \bibinfo{person}{Rongzi
  Ji}.} \bibinfo{year}{2013}\natexlab{}.
\newblock \showarticletitle{Label Distribution Learning}. In
  \bibinfo{booktitle}{\emph{2013 IEEE 13th International Conference on Data
  Mining Workshops}}. IEEE, \bibinfo{pages}{377--383}.
\newblock


\bibitem[Geng et~al\mbox{.}(2010)]%
        {geng2010facial}
\bibfield{author}{\bibinfo{person}{Xin Geng}, \bibinfo{person}{Kate
  Smith-Miles}, {and} \bibinfo{person}{Zhi-Hua Zhou}.}
  \bibinfo{year}{2010}\natexlab{}.
\newblock \showarticletitle{Facial Age Estimation by Learning from Label
  Distributions}. In \bibinfo{booktitle}{\emph{Proceedings of the AAAI
  Conference on Artificial Intelligence}}, Vol.~\bibinfo{volume}{24}.
  \bibinfo{pages}{451--456}.
\newblock


\bibitem[Geng et~al\mbox{.}(2014)]%
        {geng2014facial}
\bibfield{author}{\bibinfo{person}{Xin Geng}, \bibinfo{person}{Qin Wang}, {and}
  \bibinfo{person}{Yu Xia}.} \bibinfo{year}{2014}\natexlab{}.
\newblock \showarticletitle{Facial age estimation by adaptive label
  distribution learning}. In \bibinfo{booktitle}{\emph{2014 22nd International
  Conference on Pattern Recognition}}. IEEE, \bibinfo{pages}{4465--4470}.
\newblock


\bibitem[Geng et~al\mbox{.}(2013)]%
        {geng2013facial}
\bibfield{author}{\bibinfo{person}{Xin Geng}, \bibinfo{person}{Chao Yin}, {and}
  \bibinfo{person}{Zhi-Hua Zhou}.} \bibinfo{year}{2013}\natexlab{}.
\newblock \showarticletitle{Facial age estimation by learning from label
  distributions}.
\newblock \bibinfo{journal}{\emph{IEEE Transactions on Pattern Analysis and
  Machine Intelligence}} \bibinfo{volume}{35}, \bibinfo{number}{10}
  (\bibinfo{year}{2013}), \bibinfo{pages}{2401--2412}.
\newblock


\bibitem[Hsu et~al\mbox{.}(2017)]%
        {hsu2017automatic}
\bibfield{author}{\bibinfo{person}{Yu-Liang Hsu}, \bibinfo{person}{Jeen-Shing
  Wang}, \bibinfo{person}{Wei-Chun Chiang}, {and} \bibinfo{person}{Chien-Han
  Hung}.} \bibinfo{year}{2017}\natexlab{}.
\newblock \showarticletitle{Automatic ECG-based emotion recognition in music
  listening}.
\newblock \bibinfo{journal}{\emph{IEEE Transactions on Affective Computing}}
  \bibinfo{volume}{11}, \bibinfo{number}{1} (\bibinfo{year}{2017}),
  \bibinfo{pages}{85--99}.
\newblock


\bibitem[Jia et~al\mbox{.}(2023)]%
        {jia2023label}
\bibfield{author}{\bibinfo{person}{Xiuyi Jia}, \bibinfo{person}{Xiaoxia Shen},
  \bibinfo{person}{Weiwei Li}, \bibinfo{person}{Yunan Lu}, {and}
  \bibinfo{person}{Jihua Zhu}.} \bibinfo{year}{2023}\natexlab{}.
\newblock \showarticletitle{Label Distribution Learning by Maintaining Label
  Ranking Relation}.
\newblock \bibinfo{journal}{\emph{IEEE Transactions on Knowledge \& Data
  Engineering}} \bibinfo{volume}{35}, \bibinfo{number}{02}
  (\bibinfo{year}{2023}), \bibinfo{pages}{1695--1707}.
\newblock


\bibitem[Jia et~al\mbox{.}(2019)]%
        {jia2019facial}
\bibfield{author}{\bibinfo{person}{Xiuyi Jia}, \bibinfo{person}{Xiang Zheng},
  \bibinfo{person}{Weiwei Li}, \bibinfo{person}{Changqing Zhang}, {and}
  \bibinfo{person}{Zechao Li}.} \bibinfo{year}{2019}\natexlab{}.
\newblock \showarticletitle{Facial emotion distribution learning by exploiting
  low-rank label correlations locally}. In
  \bibinfo{booktitle}{\emph{Proceedings of the IEEE/CVF Conference on Computer
  Vision and Pattern Recognition}}. \bibinfo{pages}{9841--9850}.
\newblock


\bibitem[Jiang et~al\mbox{.}(2023)]%
        {jiang2023multimodal}
\bibfield{author}{\bibinfo{person}{Wei-Bang Jiang}, \bibinfo{person}{Xuan-Hao
  Liu}, \bibinfo{person}{Wei-Long Zheng}, {and} \bibinfo{person}{Bao-Liang
  Lu}.} \bibinfo{year}{2023}\natexlab{}.
\newblock \showarticletitle{Multimodal adaptive emotion transformer with
  flexible modality inputs on a novel dataset with continuous labels}. In
  \bibinfo{booktitle}{\emph{Proceedings of the 31st ACM International
  Conference on Multimedia}}. \bibinfo{pages}{5975--5984}.
\newblock


\bibitem[Kantorovich(2006)]%
        {kantorovich2006translocation}
\bibfield{author}{\bibinfo{person}{Leonid~V Kantorovich}.}
  \bibinfo{year}{2006}\natexlab{}.
\newblock \showarticletitle{On the Translocation of Masses.}
\newblock \bibinfo{journal}{\emph{Journal of Mathematical Sciences}}
  \bibinfo{volume}{133}, \bibinfo{number}{4} (\bibinfo{year}{2006}).
\newblock


\bibitem[Kim and Andr{\'e}(2008)]%
        {kim2008emotion}
\bibfield{author}{\bibinfo{person}{Jonghwa Kim} {and}
  \bibinfo{person}{Elisabeth Andr{\'e}}.} \bibinfo{year}{2008}\natexlab{}.
\newblock \showarticletitle{Emotion recognition based on physiological changes
  in music listening}.
\newblock \bibinfo{journal}{\emph{IEEE transactions on pattern analysis and
  machine intelligence}} \bibinfo{volume}{30}, \bibinfo{number}{12}
  (\bibinfo{year}{2008}), \bibinfo{pages}{2067--2083}.
\newblock


\bibitem[Koelstra et~al\mbox{.}(2011)]%
        {koelstra2011deap}
\bibfield{author}{\bibinfo{person}{Sander Koelstra}, \bibinfo{person}{Christian
  Muhl}, \bibinfo{person}{Mohammad Soleymani}, \bibinfo{person}{Jong-Seok Lee},
  \bibinfo{person}{Ashkan Yazdani}, \bibinfo{person}{Touradj Ebrahimi},
  \bibinfo{person}{Thierry Pun}, \bibinfo{person}{Anton Nijholt}, {and}
  \bibinfo{person}{Ioannis Patras}.} \bibinfo{year}{2011}\natexlab{}.
\newblock \showarticletitle{Deap: A database for emotion analysis; using
  physiological signals}.
\newblock \bibinfo{journal}{\emph{IEEE transactions on affective computing}}
  \bibinfo{volume}{3}, \bibinfo{number}{1} (\bibinfo{year}{2011}),
  \bibinfo{pages}{18--31}.
\newblock


\bibitem[Kou et~al\mbox{.}(2024)]%
        {ijcai2024p478}
\bibfield{author}{\bibinfo{person}{Zhiqiang Kou}, \bibinfo{person}{Jing Wang},
  \bibinfo{person}{Jiawei Tang}, \bibinfo{person}{Yuheng Jia},
  \bibinfo{person}{Boyu Shi}, {and} \bibinfo{person}{Xin Geng}.}
  \bibinfo{year}{2024}\natexlab{}.
\newblock \showarticletitle{Exploiting Multi-Label Correlation in Label
  Distribution Learning}. In \bibinfo{booktitle}{\emph{Proceedings of the
  Thirty-Third International Joint Conference on Artificial Intelligence,
  {IJCAI-24}}}, \bibfield{editor}{\bibinfo{person}{Kate Larson}} (Ed.).
  \bibinfo{publisher}{International Joint Conferences on Artificial
  Intelligence Organization}, \bibinfo{pages}{4326--4334}.
\newblock
\href{https://doi.org/10.24963/ijcai.2024/478}{doi:\nolinkurl{10.24963/ijcai.2024/478}}
\newblock
\shownote{Main Track}.


\bibitem[Lang(1995)]%
        {lang1995emotion}
\bibfield{author}{\bibinfo{person}{Peter~J Lang}.}
  \bibinfo{year}{1995}\natexlab{}.
\newblock \showarticletitle{The emotion probe: Studies of motivation and
  attention.}
\newblock \bibinfo{journal}{\emph{American psychologist}} \bibinfo{volume}{50},
  \bibinfo{number}{5} (\bibinfo{year}{1995}), \bibinfo{pages}{372}.
\newblock


\bibitem[Li et~al\mbox{.}(2023)]%
        {li2023effective}
\bibfield{author}{\bibinfo{person}{Cunbo Li}, \bibinfo{person}{Peiyang Li},
  \bibinfo{person}{Yangsong Zhang}, \bibinfo{person}{Ning Li},
  \bibinfo{person}{Yajing Si}, \bibinfo{person}{Fali Li},
  \bibinfo{person}{Zehong Cao}, \bibinfo{person}{Huafu Chen},
  \bibinfo{person}{Badong Chen}, \bibinfo{person}{Dezhong Yao},
  {et~al\mbox{.}}} \bibinfo{year}{2023}\natexlab{}.
\newblock \showarticletitle{Effective emotion recognition by learning
  discriminative graph topologies in EEG brain networks}.
\newblock \bibinfo{journal}{\emph{IEEE Transactions on Neural Networks and
  Learning Systems}} (\bibinfo{year}{2023}).
\newblock


\bibitem[Li and Deng(2019)]%
        {li2019blended}
\bibfield{author}{\bibinfo{person}{Shan Li} {and} \bibinfo{person}{Weihong
  Deng}.} \bibinfo{year}{2019}\natexlab{}.
\newblock \showarticletitle{Blended emotion in-the-wild: Multi-label facial
  expression recognition using crowdsourced annotations and deep locality
  feature learning}.
\newblock \bibinfo{journal}{\emph{International Journal of Computer Vision}}
  \bibinfo{volume}{127}, \bibinfo{number}{6} (\bibinfo{year}{2019}),
  \bibinfo{pages}{884--906}.
\newblock


\bibitem[Liu et~al\mbox{.}(2023)]%
        {liu2023emotion}
\bibfield{author}{\bibinfo{person}{Fang Liu}, \bibinfo{person}{Pei Yang},
  \bibinfo{person}{Yezhi Shu}, \bibinfo{person}{Fei Yan},
  \bibinfo{person}{Guanhua Zhang}, {and} \bibinfo{person}{Yong-Jin Liu}.}
  \bibinfo{year}{2023}\natexlab{}.
\newblock \showarticletitle{Emotion dictionary learning with modality
  attentions for mixed emotion exploration}.
\newblock \bibinfo{journal}{\emph{IEEE Transactions on Affective Computing}}
  (\bibinfo{year}{2023}).
\newblock


\bibitem[Liu et~al\mbox{.}(2025)]%
        {10930808}
\bibfield{author}{\bibinfo{person}{Minxu Liu}, \bibinfo{person}{Donghai Guan},
  \bibinfo{person}{Chuhang Zheng}, {and} \bibinfo{person}{Qi Zhu}.}
  \bibinfo{year}{2025}\natexlab{}.
\newblock \showarticletitle{Multi-Modal Discriminative Network for Emotion
  Recognition across Individuals}.
\newblock \bibinfo{journal}{\emph{IEEE Transactions on Cognitive and
  Developmental Systems}} (\bibinfo{year}{2025}), \bibinfo{pages}{1--13}.
\newblock
\href{https://doi.org/10.1109/TCDS.2025.3552124}{doi:\nolinkurl{10.1109/TCDS.2025.3552124}}


\bibitem[Ngai et~al\mbox{.}(2022)]%
        {ngai2022emotion}
\bibfield{author}{\bibinfo{person}{Wang~Kay Ngai}, \bibinfo{person}{Haoran
  Xie}, \bibinfo{person}{Di Zou}, {and} \bibinfo{person}{Kee-Lee Chou}.}
  \bibinfo{year}{2022}\natexlab{}.
\newblock \showarticletitle{Emotion recognition based on convolutional neural
  networks and heterogeneous bio-signal data sources}.
\newblock \bibinfo{journal}{\emph{Information Fusion}}  \bibinfo{volume}{77}
  (\bibinfo{year}{2022}), \bibinfo{pages}{107--117}.
\newblock


\bibitem[Peng et~al\mbox{.}(2024)]%
        {peng2024carat}
\bibfield{author}{\bibinfo{person}{Cheng Peng}, \bibinfo{person}{Ke Chen},
  \bibinfo{person}{Lidan Shou}, {and} \bibinfo{person}{Gang Chen}.}
  \bibinfo{year}{2024}\natexlab{}.
\newblock \showarticletitle{CARAT: Contrastive Feature Reconstruction and
  Aggregation for Multi-Modal Multi-Label Emotion Recognition}. In
  \bibinfo{booktitle}{\emph{Proceedings of the AAAI Conference on Artificial
  Intelligence}}, Vol.~\bibinfo{volume}{38}. \bibinfo{pages}{14581--14589}.
\newblock


\bibitem[Perrot et~al\mbox{.}(2016)]%
        {perrot2016mapping}
\bibfield{author}{\bibinfo{person}{Micha{\"e}l Perrot},
  \bibinfo{person}{Nicolas Courty}, \bibinfo{person}{R{\'e}mi Flamary}, {and}
  \bibinfo{person}{Amaury Habrard}.} \bibinfo{year}{2016}\natexlab{}.
\newblock \showarticletitle{Mapping estimation for discrete optimal transport}.
\newblock \bibinfo{journal}{\emph{Advances in Neural Information Processing
  Systems}}  \bibinfo{volume}{29} (\bibinfo{year}{2016}).
\newblock


\bibitem[Schmidt et~al\mbox{.}(2018)]%
        {schmidt2018introducing}
\bibfield{author}{\bibinfo{person}{Philip Schmidt}, \bibinfo{person}{Attila
  Reiss}, \bibinfo{person}{Robert Duerichen}, \bibinfo{person}{Claus
  Marberger}, {and} \bibinfo{person}{Kristof Van~Laerhoven}.}
  \bibinfo{year}{2018}\natexlab{}.
\newblock \showarticletitle{Introducing wesad, a multimodal dataset for
  wearable stress and affect detection}. In
  \bibinfo{booktitle}{\emph{Proceedings of the 20th ACM International
  Conference on Multimodal Interaction}}. \bibinfo{pages}{400--408}.
\newblock


\bibitem[Shneiderman and Plaisant(2010)]%
        {shneiderman2010designing}
\bibfield{author}{\bibinfo{person}{Ben Shneiderman} {and}
  \bibinfo{person}{Catherine Plaisant}.} \bibinfo{year}{2010}\natexlab{}.
\newblock \bibinfo{booktitle}{\emph{Designing the user interface: strategies
  for effective human-computer interaction}}.
\newblock \bibinfo{publisher}{Pearson Education India}.
\newblock


\bibitem[Shu et~al\mbox{.}(2022)]%
        {shu2022emotion}
\bibfield{author}{\bibinfo{person}{Yezhi Shu}, \bibinfo{person}{Pei Yang},
  \bibinfo{person}{Niqi Liu}, \bibinfo{person}{Shu Zhang},
  \bibinfo{person}{Guozhen Zhao}, {and} \bibinfo{person}{Yong-Jin Liu}.}
  \bibinfo{year}{2022}\natexlab{}.
\newblock \showarticletitle{Emotion distribution learning based on peripheral
  physiological signals}.
\newblock \bibinfo{journal}{\emph{IEEE Transactions on Affective Computing}}
  \bibinfo{volume}{14}, \bibinfo{number}{3} (\bibinfo{year}{2022}),
  \bibinfo{pages}{2470--2483}.
\newblock


\bibitem[Song et~al\mbox{.}(2024)]%
        {song2024multimodal}
\bibfield{author}{\bibinfo{person}{Andrew~H Song}, \bibinfo{person}{Richard~J
  Chen}, \bibinfo{person}{Guillaume Jaume}, \bibinfo{person}{Anurag~J Vaidya},
  \bibinfo{person}{Alexander~S Baras}, {and} \bibinfo{person}{Faisal Mahmood}.}
  \bibinfo{year}{2024}\natexlab{}.
\newblock \showarticletitle{Multimodal Prototyping for cancer survival
  prediction}.
\newblock \bibinfo{journal}{\emph{arXiv preprint arXiv:2407.00224}}
  (\bibinfo{year}{2024}).
\newblock


\bibitem[Song et~al\mbox{.}(2022)]%
        {song2022eeg}
\bibfield{author}{\bibinfo{person}{Yonghao Song}, \bibinfo{person}{Qingqing
  Zheng}, \bibinfo{person}{Bingchuan Liu}, {and} \bibinfo{person}{Xiaorong
  Gao}.} \bibinfo{year}{2022}\natexlab{}.
\newblock \showarticletitle{EEG conformer: Convolutional transformer for EEG
  decoding and visualization}.
\newblock \bibinfo{journal}{\emph{IEEE Transactions on Neural Systems and
  Rehabilitation Engineering}}  \bibinfo{volume}{31} (\bibinfo{year}{2022}),
  \bibinfo{pages}{710--719}.
\newblock


\bibitem[Udovi{\v{c}}i{\'c} et~al\mbox{.}(2017)]%
        {udovivcic2017wearable}
\bibfield{author}{\bibinfo{person}{Goran Udovi{\v{c}}i{\'c}},
  \bibinfo{person}{Jurica {\DH}erek}, \bibinfo{person}{Mladen Russo}, {and}
  \bibinfo{person}{Marjan Sikora}.} \bibinfo{year}{2017}\natexlab{}.
\newblock \showarticletitle{Wearable emotion recognition system based on GSR
  and PPG signals}. In \bibinfo{booktitle}{\emph{Proceedings of the 2nd
  international workshop on multimedia for personal health and health care}}.
  \bibinfo{pages}{53--59}.
\newblock


\bibitem[Van~der Maaten and Hinton(2008)]%
        {van2008visualizing}
\bibfield{author}{\bibinfo{person}{Laurens Van~der Maaten} {and}
  \bibinfo{person}{Geoffrey Hinton}.} \bibinfo{year}{2008}\natexlab{}.
\newblock \showarticletitle{Visualizing data using t-SNE.}
\newblock \bibinfo{journal}{\emph{Journal of Machine Learning Research}}
  \bibinfo{volume}{9}, \bibinfo{number}{11} (\bibinfo{year}{2008}).
\newblock


\bibitem[Verma et~al\mbox{.}(2021)]%
        {verma2021automer}
\bibfield{author}{\bibinfo{person}{Monu Verma}, \bibinfo{person}{M~Satish~Kumar
  Reddy}, \bibinfo{person}{Yashwanth~Reddy Meedimale}, \bibinfo{person}{Murari
  Mandal}, {and} \bibinfo{person}{Santosh~Kumar Vipparthi}.}
  \bibinfo{year}{2021}\natexlab{}.
\newblock \showarticletitle{Automer: Spatiotemporal neural architecture search
  for microexpression recognition}.
\newblock \bibinfo{journal}{\emph{IEEE Transactions on Neural Networks and
  Learning Systems}} \bibinfo{volume}{33}, \bibinfo{number}{11}
  (\bibinfo{year}{2021}), \bibinfo{pages}{6116--6128}.
\newblock


\bibitem[Watson(1994)]%
        {watson1994panas}
\bibfield{author}{\bibinfo{person}{D Watson}.} \bibinfo{year}{1994}\natexlab{}.
\newblock \showarticletitle{The PANAS-X: Manual for the positive and negative
  affect schedule-expanded form}.
\newblock \bibinfo{journal}{\emph{The University of Iowa}}
  (\bibinfo{year}{1994}).
\newblock


\bibitem[Wen et~al\mbox{.}(2023)]%
        {wen2023ordinal}
\bibfield{author}{\bibinfo{person}{Changsong Wen}, \bibinfo{person}{Xin Zhang},
  \bibinfo{person}{Xingxu Yao}, {and} \bibinfo{person}{Jufeng Yang}.}
  \bibinfo{year}{2023}\natexlab{}.
\newblock \showarticletitle{Ordinal label distribution learning}. In
  \bibinfo{booktitle}{\emph{Proceedings of the IEEE/CVF International
  Conference on Computer Vision}}. \bibinfo{pages}{23481--23491}.
\newblock


\bibitem[Williams and Aaker(2002)]%
        {williams2002can}
\bibfield{author}{\bibinfo{person}{Patti Williams} {and}
  \bibinfo{person}{Jennifer~L Aaker}.} \bibinfo{year}{2002}\natexlab{}.
\newblock \showarticletitle{Can mixed emotions peacefully coexist?}
\newblock \bibinfo{journal}{\emph{Journal of consumer research}}
  \bibinfo{volume}{28}, \bibinfo{number}{4} (\bibinfo{year}{2002}),
  \bibinfo{pages}{636--649}.
\newblock


\bibitem[Yan et~al\mbox{.}(2014)]%
        {yan2014casme}
\bibfield{author}{\bibinfo{person}{Wen-Jing Yan}, \bibinfo{person}{Xiaobai Li},
  \bibinfo{person}{Su-Jing Wang}, \bibinfo{person}{Guoying Zhao},
  \bibinfo{person}{Yong-Jin Liu}, \bibinfo{person}{Yu-Hsin Chen}, {and}
  \bibinfo{person}{Xiaolan Fu}.} \bibinfo{year}{2014}\natexlab{}.
\newblock \showarticletitle{CASME II: An improved spontaneous micro-expression
  database and the baseline evaluation}.
\newblock \bibinfo{journal}{\emph{PloS one}} \bibinfo{volume}{9},
  \bibinfo{number}{1} (\bibinfo{year}{2014}), \bibinfo{pages}{e86041}.
\newblock


\bibitem[Yang et~al\mbox{.}(2024)]%
        {yang2024multimodal}
\bibfield{author}{\bibinfo{person}{Pei Yang}, \bibinfo{person}{Niqi Liu},
  \bibinfo{person}{Xinge Liu}, \bibinfo{person}{Yezhi Shu},
  \bibinfo{person}{Wenqi Ji}, \bibinfo{person}{Ziqi Ren},
  \bibinfo{person}{Jenny Sheng}, \bibinfo{person}{Minjing Yu},
  \bibinfo{person}{Ran Yi}, \bibinfo{person}{Dan Zhang}, {et~al\mbox{.}}}
  \bibinfo{year}{2024}\natexlab{}.
\newblock \showarticletitle{A Multimodal Dataset for Mixed Emotion
  Recognition}.
\newblock \bibinfo{journal}{\emph{Scientific Data}} \bibinfo{volume}{11},
  \bibinfo{number}{1} (\bibinfo{year}{2024}), \bibinfo{pages}{847}.
\newblock


\bibitem[Ye et~al\mbox{.}(2022)]%
        {ye2022hierarchical}
\bibfield{author}{\bibinfo{person}{Mengqing Ye}, \bibinfo{person}{CL~Philip
  Chen}, {and} \bibinfo{person}{Tong Zhang}.} \bibinfo{year}{2022}\natexlab{}.
\newblock \showarticletitle{Hierarchical dynamic graph convolutional network
  with interpretability for EEG-based emotion recognition}.
\newblock \bibinfo{journal}{\emph{IEEE Transactions on Neural Networks and
  Learning Systems}} (\bibinfo{year}{2022}).
\newblock


\bibitem[Yi and Mak(2020)]%
        {yi2020improving}
\bibfield{author}{\bibinfo{person}{Lu Yi} {and} \bibinfo{person}{Man-Wai Mak}.}
  \bibinfo{year}{2020}\natexlab{}.
\newblock \showarticletitle{Improving speech emotion recognition with
  adversarial data augmentation network}.
\newblock \bibinfo{journal}{\emph{IEEE Transactions on Neural Networks and
  Learning Systems}} \bibinfo{volume}{33}, \bibinfo{number}{1}
  (\bibinfo{year}{2020}), \bibinfo{pages}{172--184}.
\newblock


\bibitem[Zhang et~al\mbox{.}(2020)]%
        {zhang2020multi}
\bibfield{author}{\bibinfo{person}{Dong Zhang}, \bibinfo{person}{Xincheng Ju},
  \bibinfo{person}{Junhui Li}, \bibinfo{person}{Shoushan Li},
  \bibinfo{person}{Qiaoming Zhu}, {and} \bibinfo{person}{Guodong Zhou}.}
  \bibinfo{year}{2020}\natexlab{}.
\newblock \showarticletitle{Multi-modal multi-label emotion detection with
  modality and label dependence}. In \bibinfo{booktitle}{\emph{Proceedings of
  the 2020 Conference on Empirical Methods in Natural Language processing
  (EMNLP)}}. \bibinfo{pages}{3584--3593}.
\newblock


\bibitem[Zhang et~al\mbox{.}(2021a)]%
        {zhang2021multi}
\bibfield{author}{\bibinfo{person}{Dong Zhang}, \bibinfo{person}{Xincheng Ju},
  \bibinfo{person}{Wei Zhang}, \bibinfo{person}{Junhui Li},
  \bibinfo{person}{Shoushan Li}, \bibinfo{person}{Qiaoming Zhu}, {and}
  \bibinfo{person}{Guodong Zhou}.} \bibinfo{year}{2021}\natexlab{a}.
\newblock \showarticletitle{Multi-modal multi-label emotion recognition with
  heterogeneous hierarchical message passing}. In
  \bibinfo{booktitle}{\emph{Proceedings of the AAAI Conference on Artificial
  Intelligence}}, Vol.~\bibinfo{volume}{35}. \bibinfo{pages}{14338--14346}.
\newblock


\bibitem[Zhang et~al\mbox{.}(2018)]%
        {zhang2018text}
\bibfield{author}{\bibinfo{person}{Yuxiang Zhang}, \bibinfo{person}{Jiamei Fu},
  \bibinfo{person}{Dongyu She}, \bibinfo{person}{Ying Zhang},
  \bibinfo{person}{Senzhang Wang}, {and} \bibinfo{person}{Jufeng Yang}.}
  \bibinfo{year}{2018}\natexlab{}.
\newblock \showarticletitle{Text Emotion Distribution Learning via Multi-Task
  Convolutional Neural Network.}. In \bibinfo{booktitle}{\emph{IJCAI}}.
  \bibinfo{pages}{4595--4601}.
\newblock


\bibitem[Zhang et~al\mbox{.}(2021b)]%
        {zhang2021cped}
\bibfield{author}{\bibinfo{person}{Yulin Zhang}, \bibinfo{person}{Guozhen
  Zhao}, \bibinfo{person}{Yezhi Shu}, \bibinfo{person}{Yan Ge},
  \bibinfo{person}{Dan Zhang}, \bibinfo{person}{Yong-Jin Liu}, {and}
  \bibinfo{person}{Xianghong Sun}.} \bibinfo{year}{2021}\natexlab{b}.
\newblock \showarticletitle{CPED: A Chinese positive emotion database for
  emotion elicitation and analysis}.
\newblock \bibinfo{journal}{\emph{IEEE Transactions on Affective Computing}}
  \bibinfo{volume}{14}, \bibinfo{number}{2} (\bibinfo{year}{2021}),
  \bibinfo{pages}{1417--1430}.
\newblock


\bibitem[Zhao and Pietikainen(2007)]%
        {zhao2007dynamic}
\bibfield{author}{\bibinfo{person}{Guoying Zhao} {and} \bibinfo{person}{Matti
  Pietikainen}.} \bibinfo{year}{2007}\natexlab{}.
\newblock \showarticletitle{Dynamic texture recognition using local binary
  patterns with an application to facial expressions}.
\newblock \bibinfo{journal}{\emph{IEEE transactions on pattern analysis and
  machine intelligence}} \bibinfo{volume}{29}, \bibinfo{number}{6}
  (\bibinfo{year}{2007}), \bibinfo{pages}{915--928}.
\newblock


\bibitem[Zhao et~al\mbox{.}(2020)]%
        {zhao2020multi}
\bibfield{author}{\bibinfo{person}{Guozhen Zhao}, \bibinfo{person}{Yulin
  Zhang}, \bibinfo{person}{Guanhua Zhang}, \bibinfo{person}{Dan Zhang}, {and}
  \bibinfo{person}{Yong-Jin Liu}.} \bibinfo{year}{2020}\natexlab{}.
\newblock \showarticletitle{Multi-target positive emotion recognition from EEG
  signals}.
\newblock \bibinfo{journal}{\emph{IEEE Transactions on Affective Computing}}
  \bibinfo{volume}{14}, \bibinfo{number}{1} (\bibinfo{year}{2020}),
  \bibinfo{pages}{370--381}.
\newblock


\bibitem[Zheng et~al\mbox{.}(2023)]%
        {zheng2023prior}
\bibfield{author}{\bibinfo{person}{Chuhang Zheng}, \bibinfo{person}{Wei Shao},
  \bibinfo{person}{Daoqiang Zhang}, {and} \bibinfo{person}{Qi Zhu}.}
  \bibinfo{year}{2023}\natexlab{}.
\newblock \showarticletitle{Prior-driven dynamic brain networks for multi-modal
  emotion recognition}. In \bibinfo{booktitle}{\emph{International Conference
  on Medical Image Computing and Computer-Assisted Intervention}}. Springer,
  \bibinfo{pages}{389--398}.
\newblock


\bibitem[Zheng and Lu(2016)]%
        {zheng2016personalizing}
\bibfield{author}{\bibinfo{person}{Wei-Long Zheng} {and}
  \bibinfo{person}{Bao-Liang Lu}.} \bibinfo{year}{2016}\natexlab{}.
\newblock \showarticletitle{Personalizing EEG-based affective models with
  transfer learning}. In \bibinfo{booktitle}{\emph{Proceedings of the
  twenty-fifth International Joint Conference on Artificial Intelligence}}.
  \bibinfo{pages}{2732--2738}.
\newblock


\bibitem[Zhu et~al\mbox{.}(2024)]%
        {10349925}
\bibfield{author}{\bibinfo{person}{Qi Zhu}, \bibinfo{person}{Chuhang Zheng},
  \bibinfo{person}{Zheng Zhang}, \bibinfo{person}{Wei Shao}, {and}
  \bibinfo{person}{Daoqiang Zhang}.} \bibinfo{year}{2024}\natexlab{}.
\newblock \showarticletitle{Dynamic Confidence-Aware Multi-Modal Emotion
  Recognition}.
\newblock \bibinfo{journal}{\emph{IEEE Transactions on Affective Computing}}
  \bibinfo{volume}{15}, \bibinfo{number}{3} (\bibinfo{year}{2024}),
  \bibinfo{pages}{1358--1370}.
\newblock
\href{https://doi.org/10.1109/TAFFC.2023.3340924}{doi:\nolinkurl{10.1109/TAFFC.2023.3340924}}


\bibitem[Zhu et~al\mbox{.}(2025)]%
        {10938180}
\bibfield{author}{\bibinfo{person}{Qi Zhu}, \bibinfo{person}{Ting Zhu},
  \bibinfo{person}{Lunke Fei}, \bibinfo{person}{Chuhang Zheng},
  \bibinfo{person}{Wei Shao}, \bibinfo{person}{David Zhang}, {and}
  \bibinfo{person}{Daoqiang Zhang}.} \bibinfo{year}{2025}\natexlab{}.
\newblock \showarticletitle{Multi-Modal Cross-Subject Emotion Feature Alignment
  and Recognition with EEG and Eye Movements}.
\newblock \bibinfo{journal}{\emph{IEEE Transactions on Affective Computing}}
  (\bibinfo{year}{2025}), \bibinfo{pages}{1--15}.
\newblock
\href{https://doi.org/10.1109/TAFFC.2025.3554399}{doi:\nolinkurl{10.1109/TAFFC.2025.3554399}}


\end{thebibliography}
\appendix

\clearpage
\setcounter{page}{1}
\def\maketitlesupplementary
   {
   \newpage
       \twocolumn[
        \centering
        \Large
        \textbf{HeLo: Heterogeneous Multi-Modal Fusion with Label Correlation for Emotion Distribution Learning}\\
        \vspace{0.5em}Supplementary Material \\
        \vspace{1.0em}
       ] 
   }
\maketitlesupplementary

\section{Dataset Preprocessing}

\subsection{DMER Dataset}
In the DMER dataset, we follow the preprocessing pipelines in the original paper~\cite{yang2024multimodal}. For EEG signals, differential entropy (DE) features in each EEG channel from 5 frequency bands ($\delta$ band (1-3Hz), $\theta$ band (4-7Hz), $\alpha$ band (8-13Hz), $\beta$ band (14-30Hz) and $\gamma$ band (31-50Hz)) were extracted. For GSR signals, the median, mean, standard deviation, minimum, maximum, ratio of minimum, and the ratio of maximum of the raw GSR signal and its first-order and second-order derivatives were extracted as time domain statistical features according to the feature extraction~\cite{udovivcic2017wearable}. Then, the Fast Fourier Transform (FFT) was adopted to transform the GSR signals from the time domain into the frequency domain, and the median, mean, standard deviation, maximum, minimum, and range of the signal were extracted~\cite{udovivcic2017wearable}. In addition, the power spectral density (PSD) of the frequency band [0, 2] Hz was extracted using Welch’s power spectral density. The same features except PSD were extracted for the PPG signals. Notably, the size of the sliding window for both GSR and PPG feature extraction was 5 seconds with an 80\% overlap between two consecutive windows~\cite{zhang2021cped}. For facial video data, local binary patterns from three orthogonal planes (LBP-TOP) \cite{zhao2007dynamic,yan2014casme} were used for the feature extraction. Due to the missing data in subject 3, 4, 13, 16, 17, 27, and 31, the data of the remaining 73 subjects are chosen for further experiments. After all the preprocessing steps, a 913-dimension feature vector, including 90 EEG features, 28 GSR features, 27 PPG features, and 768 video features for a specific subject was obtained for further experiments. The summary of extracted features is listed in Table \ref{tabappendix:tab1}.
\subsection{WESAD Dataset}
In the WESAD dataset, features were computed with a window size of five seconds. The feature extraction methods refer to \cite{koelstra2011deap,schmidt2018introducing}. For ACC data, the mean, and standard deviation for each axis were separately and summed over all axes, absolute integral for each/all axis, and peak frequency for each axis were extracted. Following~\cite{kim2008emotion}, the heartbeats were accurately localized in ECG signals (R-peaks) to calculate the inter beat intervals (IBI). Using IBI values, the heart rate (HR) and heart rate variability (HRV) time series were calculated. For EDA data, the mean and standard deviation of the signal, the number of skin conductance response (SCR) peaks, and the mean amplitude of these peaks were extracted. For EMG data, the mean, standard deviation, range, integral, and median of the signal were extracted, along with the 10th and 90th percentiles. In the frequency domain, the peak frequency and total power spectral density were computed. Additionally, the number of peaks, mean peak amplitude, the standard deviation of peak amplitudes, the sum of peak amplitudes, and the normalized sum of peak amplitudes were calculated. Due to sensor malfunction in subject 1, and 12 and missing labels in subject 5, the data of the remaining 14 participants was used for experiments. After all the preprocessing steps, we obtained 12 ACC features, 73 ECG features, 4 EDA features, and 14 EMG features. The summary of extracted features is illustrated in Table \ref{tabappendix:tab2}.

\begin{table}[]
\caption{Extracted Affective Features for each Modality (feature dimension stated in parenthesis) on the DMER dataset.}
\resizebox{\linewidth}{!}{
\begin{tabular}{cl}
\toprule
\textbf{Modality}      & \multicolumn{1}{c}{\textbf{Extracted features}}                                                                                                                                                                                                                                                                                                                                                            \\ \midrule
EEG (90)      & \begin{tabular}[c]{@{}l@{}}5 bands (sigma, theta, alpha,   beta, gamma) differential \\ entropy (DE) features for each EEG channel.\end{tabular}                                                                                                                                                                                                                                                  \\ \midrule
GSR (28)      & \begin{tabular}[c]{@{}l@{}}\textbf{Time domain:} median, mean, standard deviation, minimum, \\ maximum, ratio of minimum, ratio of maximum of the raw \\ GSR signal, first-order and second-order derivatives. \\ \textbf{Frequency domain:} fast fourier transform (FFT), median, \\ mean, standard   deviation, maximum, minimum, range of \\ the signal, PSD of frequency band {[}0,   2{]} Hz.\end{tabular} \\ \midrule
PPG (27)      & \begin{tabular}[c]{@{}l@{}}\textbf{Time domain:} median, mean, standard deviation, minimum, \\ maximum, ratio of minimum, ratio of maximum of the raw \\ GSR signal, first-order and second-order derivatives. \\ \textbf{Frequency domain:} fast fourier transform (FFT), median, \\ mean, standard deviation, maximum, minimum, range of \\ the signal.\end{tabular}                                        \\ \midrule
Video   (768) & \begin{tabular}[c]{@{}l@{}}Local binary patterns from three   orthogonal planes (LBP-TOP) \\ features from each video frame.\end{tabular}                                                                                                                                                                                                                                                         \\ \bottomrule
\end{tabular}}
\label{tabappendix:tab1}
\end{table}

\begin{table}[]
\caption{Extracted Affective Features for each Modality (feature dimension stated in parenthesis) on the WESAD dataset.}
\resizebox{\linewidth}{!}{
\begin{tabular}{cl}
\toprule
\textbf{Modality} & \multicolumn{1}{c}{\textbf{Extracted features}}                                                                                                                                                                                                                                                                                                                                      \\ \midrule
ECG (73)          & \begin{tabular}[c]{@{}l@{}}Inter beat intervals (IBI), heart rate (HR), and heart rate \\ variability (HRV) time series.\end{tabular}                                                                                                                                                                                                                                               \\ \midrule
EDA (4)           & \begin{tabular}[c]{@{}l@{}}mean/standard deviation of the signal, the number of skin \\ conductance response (SCR) peaks, the mean amplitude \\ of SCR peaks.\end{tabular}                                                                                                                                                                                                       \\ \midrule
EMG (14)          & \begin{tabular}[c]{@{}l@{}}mean, standard deviation, range,   integral, median of the \\ signal along with the 10th and 90th percentiles, peak  \\ frequency and total power spectral density, the number of \\ peaks, mean peak amplitude, standard deviation of peak \\ amplitudes, sum of peak amplitudes,   normalized sum of \\ peak amplitudes were calculated.\end{tabular} \\ \midrule
ACC (12)          & \begin{tabular}[c]{@{}l@{}}standard deviation for each axis separately and summed \\ over all axes, absolute integral for each/all axis, peak \\ frequency for each axis.\end{tabular}                                                                                                                                                                                             \\ \bottomrule
\end{tabular}}
\label{tabappendix:tab2}
\end{table}

\section{Details of the Comparison Methods}
\begin{itemize}
    \item \textbf{PT-SVM}~\cite{geng2016label} are based on Bayes and SVM classifiers with problem transformation strategy, which converts LDL to single-label distribution problems via transforming the training data into weighted single-label data.
    \item \textbf{AA-KNN}~\cite{geng2016label} is an algorithm adaptation strategy-based method, which extend classic k-Nearest Keighbor (KNN) to handle LDL problems.
    \item \textbf{SA-CPNN}~\cite{geng2013facial} is a specialized algorithm for LDL, which is built on a three-layer conditional probability neural network.
    \item \textbf{Conformer}~\cite{song2022eeg} adopts convolutional neural networks (CNNs) and Transformer architectures to mine local and global dependencies in sequential data, respectively.
    \item \textbf{MAET}~\cite{jiang2023multimodal} utilizes specialized modules for processing uni-modal or multi-modal inputs flexibly, and the subject discrepancy is alleviated by adopting adversarial training.
    \item \textbf{CARAT}~\cite{peng2024carat} adopts a reconstruction-based fusion mechanism to better model fine-grained modality-to-label dependencies by contrastively learning modal-separated and label-specific features. To further exploit the modality complementarity, a shuffle-based aggregation strategy was introduced to enrich co-occurrence collaboration among labels.
    \item \textbf{LDL-LRR}~\cite{jia2023label} adopts a novel ranking loss function for label distribution characteristics, and combines it with KL-divergence as the loss term of the objective function.
    \item \textbf{TLRLDL}~\cite{ijcai2024p478} focuses on improving label distribution learning (LDL) by leveraging low-rank label correlations specifically within an auxiliary multi-label learning framework.
    \item \textbf{CAD}~\cite{wen2023ordinal} computes the minimal cost required to transform one label distribution into another by considering these spatial relationships label-by-label.
    \item \textbf{EmotionDict}~\cite{liu2023emotion} adopts modality attention to learn cross-modal interactions and disentangle the mixed emotion representations into a weighted sum of a set of basic emotion elements in a shared latent space to enhance the emotion distribution learning performance.
\end{itemize}
Notably, the label distribution learning methods, i.e., LDL-LRR~\cite{jia2023label}, TLRLDL~\cite{ijcai2024p478}, CAD~\cite{wen2023ordinal}, is implemented using the \textbf{PyLDL} toolkit in our experiments.

\begin{table}[]
\centering
\caption{Ablation of Query Modality on the DMER dataset}
\vspace{-2mm}
\begin{tabular}{ccc>{\columncolor{lightblue}}c}
\toprule
Measure                & GSR    & PPG    & \textbf{EEG}   \\ \midrule
Cheby.-dependent (↓)   & 0.0574 & 0.0591 & \textbf{0.0446} \\
Cl.-dependent (↓)      & 0.3882 & 0.4045 & \textbf{0.3256} \\
Canb.-dependent (↓)    & 0.9219 & 0.9182 & \textbf{0.8664} \\
KL.-dependent (↓)      & 0.0446 & 0.0462 & \textbf{0.0323} \\
Cos.-dependent (↑)     & 0.9590 & 0.9498 & \textbf{0.9714} \\
Inter.-dependent (↑)   & 0.8812 & 0.8990  & \textbf{0.9128} \\ \midrule
Cheby.-independent (↓) & 0.0942 & 0.0924 & \textbf{0.0882} \\
Cl.-independent (↓)    & 0.7210 & 0.6970 & \textbf{0.6289} \\
Canb.-independent (↓)  & 1.8473 & 1.8821 & \textbf{1.7603} \\
KL.-independent (↓)    & 0.1183 & 0.1196 & \textbf{0.1027} \\
Cos.-independent (↑)   & 0.9055 & 0.9005 & \textbf{0.9148} \\
Inter.-independent (↑) & 0.7992 & 0.8026 & \textbf{0.8194} \\ \bottomrule
\end{tabular}
\label{tabappendix:tab3}
\vspace{-6mm}
\end{table}

\section{Analysis on Query Modality}
Compared with peripheral physiological signals such as GSR and PPG, EEG is more closely related to the neural mechanism of emotion generation. Using EEG as the query guides the cross-attention mechanism to focus on the emotional information at the neural level, effectively capturing the relationship between EEG and other physiological signals. We conduct ablation experiments of query modality on the DMER dataset with both subject-dependent and subject-independent settings. The ablation results were shown in Table \ref{tabappendix:tab3}, demonstrating EEG's efficacy and appropriateness for this role in our method's architecture.

\begin{table}[h]
\caption{Ablation study in modalities of subject-dependent experiment on the DMER dataset.}
\resizebox{\linewidth}{!}{
\begin{tabular}{cccc|cccccc}
\toprule
\multicolumn{4}{c|}{Modality} & \multicolumn{6}{c}{Measure}                                                                               \\ \midrule
EEG   & GSR   & PPG  & Video  & Cheby. (↓)      & Cl. (↓)         & Canb. (↓)       & KL. (↓)         & Cos. (↑)        & Inter. (↑)      \\ \midrule
\checkmark     &       &      &        & 0.0591          & 0.3722          & 0.8825          & 0.0442          & 0.9407          & 0.8896          \\
      & \checkmark     &      &        & 0.0695          & 0.4015          & 0.9210          & 0.0584          & 0.9318          & 0.8685          \\
      &       & \checkmark    &        & 0.0688          & 0.4128          & 0.9256          & 0.0559          & 0.9261          & 0.8602          \\
      &       &      & \checkmark      & 0.0602          & 0.3885          & 0.9194          & 0.0521          & 0.9335          & 0.8725          \\
\checkmark     & \checkmark     &      &        & 0.0571          & 0.3689          & 0.9112          & 0.0431          & 0.9429          & 0.8942          \\
\checkmark     &       & \checkmark    &        & 0.0580          & 0.3660          & 0.9088          & 0.0472          & 0.9446          & 0.8921          \\
\checkmark     &       &      & \checkmark      & 0.0548          & 0.3576          & 0.9011          & 0.0409          & 0.9509          & 0.8995          \\
      & \checkmark     & \checkmark    &        & 0.0595          & 0.3915          & 0.9117          & 0.0518          & 0.9445          & 0.8821          \\
      & \checkmark     &      & \checkmark      & 0.0628          & 0.3874          & 0.9050          & 0.0496          & 0.9420          & 0.8870          \\
      &       & \checkmark    & \checkmark      & 0.0609          & 0.4042          & 0.9214          & 0.0487          & 0.9392          & 0.8804          \\
\checkmark     & \checkmark     & \checkmark    &        & 0.0552          & 0.3380          & 0.9065          & 0.0377          & 0.9628          & 0.9086          \\
\checkmark     &       & \checkmark    & \checkmark      & 0.0574          & 0.3550          & 0.8987          & 0.0429          & 0.9591          & 0.9007          \\
\checkmark     & \checkmark     &      & \checkmark      & 0.0487          & 0.3671          & 0.8820          & 0.0357          & 0.9692          & 0.9086          \\
      & \checkmark     & \checkmark    & \checkmark      & 0.0529          & 0.3372          & 0.9028          & 0.0381          & 0.9677          & 0.9102          \\
\rowcolor{lightblue} \checkmark     & \checkmark     & \checkmark    & \checkmark      & \textbf{0.0446} & \textbf{0.3256} & \textbf{0.8664} & \textbf{0.0323} & \textbf{0.9714} & \textbf{0.9128} \\ \bottomrule
\end{tabular}}
\label{tabappendix:tab4}
\end{table}

\begin{table}[h]
\caption{Ablation study in modalities of subject-independent experiment on the DMER dataset.}
\resizebox{\linewidth}{!}{
\begin{tabular}{cccc|cccccc}
\toprule
\multicolumn{4}{c|}{Modality} & \multicolumn{6}{c}{Measure}                                                                               \\ \midrule
EEG   & GSR   & PPG  & Video  & Cheby. (↓)      & Cl. (↓)         & Canb. (↓)       & KL. (↓)         & Cos. (↑)        & Inter. (↑)      \\ \midrule
\checkmark     &       &      &        & 0.0991          & 0.6998          & 1.9258          & 0.1205          & 0.8901          & 0.7962          \\
      & \checkmark     &      &        & 0.1126          & 0.7218          & 1.9662          & 0.1309          & 0.8795          & 0.7889          \\
      &       & \checkmark    &        & 0.1039          & 0.7150          & 1.9581          & 0.1250          & 0.8850          & 0.7835          \\
      &       &      & \checkmark      & 0.1021          & 0.7246          & 1.9440          & 0.1276          & 0.8872          & 0.7991          \\
\checkmark     & \checkmark     &      &        & 0.0984          & 0.6762          & 1.8709          & 0.1198          & 0.9021          & 0.8083          \\
\checkmark     &       & \checkmark    &        & 0.0977          & 0.6881          & 1.8652          & 0.1186          & 0.8984          & 0.8144          \\
\checkmark     &       &      & \checkmark      & 0.0968          & 0.7017          & 1.8845          & 0.1178          & 0.8917          & 0.8098          \\
      & \checkmark     & \checkmark    &        & 0.1004          & 0.7008          & 1.9308          & 0.1210          & 0.8958          & 0.8022          \\
      & \checkmark     &      & \checkmark      & 0.0974          & 0.6925          & 1.9412          & 0.1195          & 0.8921          & 0.8035          \\
      &       & \checkmark    & \checkmark      & 0.0985          & 0.6977          & 1.9350          & 0.1226          & 0.8994          & 0.8020          \\
\checkmark     & \checkmark     & \checkmark    &        & 0.0926          & 0.6635          & 1.8250          & 0.1134          & 0.9107          & 0.8121          \\
\checkmark     &       & \checkmark    & \checkmark      & 0.0953          & 0.6742          & 1.9077          & 0.1244          & 0.9075          & 0.8056          \\
\checkmark     & \checkmark     &      & \checkmark      & 0.0964          & 0.6470          & 1.8074          & 0.1085          & 0.9122          & 0.8138          \\
      & \checkmark     & \checkmark    & \checkmark      & 0.0977          & 0.6906          & 1.9261          & 0.1175          & 0.9080          & 0.8197          \\
\rowcolor{lightblue} \checkmark     & \checkmark     & \checkmark    & \checkmark      & \textbf{0.0882} & \textbf{0.6289} & \textbf{1.7603} & \textbf{0.1027} & \textbf{0.9148} & \textbf{0.8194} \\ \bottomrule
\end{tabular}}
\label{tabappendix:tab5}
\end{table}

\begin{table}[h]
\caption{Ablation study in modalities of subject-dependent experiment on the WESAD dataset.}
\resizebox{\linewidth}{!}{
\begin{tabular}{cccc|cccccc}
\toprule
\multicolumn{4}{c|}{Modality} & \multicolumn{6}{c}{Measure}                                                                               \\ \midrule
ECG   & EMG   & EDA   & ACC   & Cheby. (↓)      & Cl. (↓)         & Canb. (↓)       & KL. (↓)         & Cos. (↑)        & Inter. (↑)      \\ \midrule
\checkmark     &       &       &       & 0.0122          & 0.0826          & 0.1729          & 0.0028          & 0.9720          & 0.9602          \\
      & \checkmark     &       &       & 0.0146          & 0.0885          & 0.1890          & 0.0042          & 0.9651          & 0.9548          \\
      &       & \checkmark     &       & 0.0137          & 0.0923          & 0.1921          & 0.0055          & 0.9622          & 0.9496          \\
      &       &       & \checkmark     & 0.0141          & 0.0848          & 0.1847          & 0.0048          & 0.9585          & 0.9447          \\
\checkmark     & \checkmark     &       &       & 0.0106          & 0.0783          & 0.1685          & 0.0023          & 0.9876          & 0.9725          \\
\checkmark     &       & \checkmark     &       & 0.0119          & 0.0790          & 0.1699          & 0.0019          & 0.9864          & 0.9748          \\
\checkmark     &       &       & \checkmark     & 0.0097          & 0.0744          & 0.1671          & 0.0026          & 0.9909          & 0.9826          \\
      & \checkmark     & \checkmark     &       & 0.0120          & 0.0831          & 0.1772          & 0.0034          & 0.9815          & 0.9705          \\
      & \checkmark     &       & \checkmark     & 0.0128          & 0.0925          & 0.1806          & 0.0035          & 0.9840          & 0.9720          \\
      &       & \checkmark     & \checkmark     & 0.0104          & 0.0887          & 0.1825          & 0.0040          & 0.9881          & 0.9736          \\
\checkmark     & \checkmark     & \checkmark     &       & 0.0082          & 0.0695          & 0.1725          & 0.0016          & 0.9971          & 0.9885          \\
\checkmark     &       & \checkmark     & \checkmark     & 0.0074          & 0.0722          & 0.1630          & 0.0016          & 0.9960          & 0.9899          \\
\checkmark     & \checkmark     &       & \checkmark     & 0.0079          & 0.0762          & 0.1658          & 0.0014          & 0.9955          & 0.9892          \\
      & \checkmark     & \checkmark     & \checkmark     & 0.0088          & 0.0744          & 0.1730          & 0.0022          & 0.9915          & 0.9873          \\
\rowcolor{lightblue} \checkmark     & \checkmark     & \checkmark     & \checkmark     & \textbf{0.0073} & \textbf{0.0653} & \textbf{0.1614} & \textbf{0.0010} & \textbf{0.9992} & \textbf{0.9905} \\ \bottomrule
\end{tabular}}
\label{tabappendix:tab6}
\end{table}

\begin{table}[h]
\caption{Ablation study in modalities of subject-independent experiment on the WESAD dataset.}
\resizebox{\linewidth}{!}{
\begin{tabular}{cccc|cccccc}
\toprule
\multicolumn{4}{c|}{Modality} & \multicolumn{6}{c}{Measure}                                                                               \\ \midrule
ECG   & EMG   & EDA   & ACC   & Cheby. (↓)      & Cl. (↓)         & Canb. (↓)       & KL. (↓)         & Cos. (↑)        & Inter. (↑)      \\ \midrule
\checkmark     &       &       &       & 0.0528          & 0.3854          & 1.0525          & 0.0389          & 0.9603          & 0.8962          \\
      & \checkmark     &       &       & 0.0539          & 0.3927          & 1.1884          & 0.0448          & 0.9518          & 0.8890          \\
      &       & \checkmark     &       & 0.0548          & 0.4018          & 1.1123          & 0.0394          & 0.9422          & 0.8748          \\
      &       &       & \checkmark     & 0.0541          & 0.3890          & 1.1091          & 0.0406          & 0.9475          & 0.8851          \\
\checkmark     & \checkmark     &       &       & 0.0520          & 0.3578          & 0.9574          & 0.0382          & 0.9642          & 0.9057          \\
\checkmark     &       & \checkmark     &       & 0.0517          & 0.3660          & 0.9971          & 0.0415          & 0.9733          & 0.9117          \\
\checkmark     &       &       & \checkmark     & 0.0510          & 0.3626          & 1.0604          & 0.0337          & 0.9715          & 0.9084          \\
      & \checkmark     & \checkmark     &       & 0.0537          & 0.3795          & 1.1072          & 0.0390          & 0.9670          & 0.8915          \\
      & \checkmark     &       & \checkmark     & 0.0518          & 0.3864          & 1.0875          & 0.0428          & 0.9506          & 0.8928          \\
      &       & \checkmark     & \checkmark     & 0.0525          & 0.3815          & 1.0256          & 0.0375          & 0.9573          & 0.9055          \\
\checkmark     & \checkmark     & \checkmark     &       & 0.0474          & 0.3842          & 1.0029          & 0.0315          & 0.9658          & 0.8991          \\
\checkmark     &       & \checkmark     & \checkmark     & 0.0419          & 0.3597          & 0.9516          & 0.0308          & 0.9722          & 0.9095          \\
\checkmark     & \checkmark     &       & \checkmark     & 0.0429          & 0.3721          & 0.9579          & 0.0344          & 0.9641          & 0.9017          \\
      & \checkmark     & \checkmark     & \checkmark     & 0.0502          & 0.3877          & 1.0748          & 0.0322          & 0.9688          & 0.8870          \\
\rowcolor{lightblue} \checkmark     & \checkmark     & \checkmark     & \checkmark     & \textbf{0.0403} & \textbf{0.3455} & \textbf{0.9329} & \textbf{0.0283} & \textbf{0.9790} & \textbf{0.9154} \\ \bottomrule
\end{tabular}}
\label{tabappendix:tab7}
\end{table}

\section{More Ablation Results on Modalities}
We conducted thorough ablation experiments on different combinations of different emotion modalities. The results are illustrated in Table \ref{tabappendix:tab4}, \ref{tabappendix:tab5}, \ref{tabappendix:tab6}, and \ref{tabappendix:tab7}. The experimental results demonstrate the necessity and effectiveness of our proposed multi-modal emotion distribution learning framework.

\begin{table}[h]
\caption{\textbf{Comparison between our method and other models used in our experiments in terms of network efficiency.}}
\resizebox{\linewidth}{!}{
\begin{tabular}{cccccc}
\toprule
\multirow{2}{*}{Metric} & TNSRE22~\cite{song2022eeg}   & ACMMM23~\cite{jiang2023multimodal} & AAAI24~\cite{peng2024carat}   & TAFFC23~\cite{liu2023emotion}     & \multirow{2}{*}{Ours} \\ \cmidrule{2-5}
                        & Conformer & MAET    & CARAT    & EmotionDict &                       \\ \midrule
FLOPs                   & 2.422M    & 91.955M & 330.671M & 156.354M    & 11.656M               \\
Params                  & 1.023M    & 15.672M & 6.780M   & 0.483M      & 4.040M                \\ \bottomrule
\end{tabular}}
\label{tabappendix:tab8}
\end{table}

\section{Model Efficiency}
We compared the efficiency of our model with 10 other methods. On one hand, deep learning-based methods generally achieve better emotion distribution learning performance than traditional label distribution learning methods. On the other hand, traditional label distribution learning typically have fewer parameters. Therefore, we conducted a comparison of FLOPs (Floating Point Operations per Second) and the number of parameters that need to be learned in the model with the other four deep learning-based methods, i.e., Conformer~\cite{song2022eeg}, MAET~\cite{jiang2023multimodal}, CARAT~\cite{peng2024carat}, and EmotionDict~\cite{liu2023emotion}). The results are summarized in Table \ref{tabappendix:tab8}.


\end{document}